\documentclass[conference]{IEEEtran}
\usepackage{cite}
\usepackage{amsmath,amssymb,amsfonts}
\usepackage{amsthm}
\usepackage{algpseudocode, algorithm}
\usepackage{graphicx}
\usepackage{textcomp}
\usepackage{orcidlink}
\usepackage{xcolor}
\usepackage{csquotes}
\usepackage{blindtext}

\definecolor{QPblue}{HTML}{003f5c}
\definecolor{Pblue}{HTML}{7aa6c2}
\definecolor{Pgreen}{HTML}{488f31}
\definecolor{Pred}{HTML}{ef5675}
\definecolor{Porange}{HTML}{ffa600}
\definecolor{Plila}{HTML}{bc5090}    
\definecolor{Wowgreen}{HTML}{009052} 

\def\BibTeX{{\rm B\kern-.05em{\sc i\kern-.025em b}\kern-.08em
    T\kern-.1667em\lower.7ex\hbox{E}\kern-.125emX}}

\usepackage[capitalize]{cleveref}
\crefname{section}{Sec.}{Secs.}
\Crefname{section}{Section}{Sections}
\crefname{table}{Tab.}{Tabs.}
\Crefname{table}{Table}{Tables}

\usepackage{bm}
\usepackage{booktabs} 

\usepackage[skip=3pt]{subcaption}

\usepackage{abbrv}
\usepackage{pgfplots}
\pgfplotsset{compat=1.18} 

\theoremstyle{plain}
\newtheorem{theorem}{Theorem}[section]

\theoremstyle{definition}
\newtheorem{definition}[theorem]{Definition}

\theoremstyle{remark}

\newcommand{\given}{\:\vert\:}
\newcommand{\matr}[1]{\mathrm{\textbf{#1}}}
\newcommand{\vect}[1]{\bm{#1}}
\newcommand{\sens}[1]{\mathrm{\Delta}_{#1}}

\newcommand{\mech}[0]{\mathcal{M}}
\newcommand{\eps}{\varepsilon}
\newcommand{\dset}{\mathcal{D}}
\newcommand{\gset}{\mathcal{G}}
\newcommand{\kset}{\mathcal{K}}
\newcommand{\stimes}{{\times}}
\newcommand{\prob}{\text{Pr}}
\newcommand{\epssel}{\eps_{\text{Sel}}}

\newcommand{\orcidcustom}[1]{%
    \hypersetup{hidelinks}%
    \orcidlink{#1}%
    \hypersetup{colorlinks=false}%
}

\begin{document}

\title{Differentially Private Active Learning:\\Balancing Effective Data Selection and Privacy}

\makeatletter 
\newcommand{\linebreakand}{%
  \end{@IEEEauthorhalign}
  \hfill\mbox{}\par
  \mbox{}\hfill\begin{@IEEEauthorhalign}
}
\makeatother 

\author{\IEEEauthorblockN{Kristian Schwethelm\IEEEauthorrefmark{1} \orcidcustom{0009-0007-4702-755X}, Johannes Kaiser\IEEEauthorrefmark{1} \orcidcustom{0009-0007-0819-8751}, Jonas Kuntzer\IEEEauthorrefmark{1} \orcidcustom{0009-0005-2903-3169}, Mehmet Yiğitsoy\IEEEauthorrefmark{5} \orcidcustom{0000-0001-6598-0933},\\Daniel Rückert\IEEEauthorrefmark{1}\IEEEauthorrefmark{2}\IEEEauthorrefmark{3} \orcidcustom{0000-0002-5683-5889}, and Georgios Kaissis\IEEEauthorrefmark{1}\IEEEauthorrefmark{3}\IEEEauthorrefmark{4} \orcidcustom{0000-0001-8382-8062}}

\IEEEauthorblockA{\IEEEauthorrefmark{1}Chair for AI in Healthcare and Medicine,
Technical University of Munich (TUM) and TUM University Hospital, Germany}
\IEEEauthorblockA{\IEEEauthorrefmark{2}Department of Computing, Imperial College London, UK}
\IEEEauthorblockA{\IEEEauthorrefmark{3}Munich Center for Machine Learning (MCML), Germany}
\IEEEauthorblockA{\IEEEauthorrefmark{4}Institute for Machine Learning in Biomedical Imaging, Helmholtz Munich, Germany}
\IEEEauthorblockA{\IEEEauthorrefmark{5}deepc GmbH, 
Munich, Germany}
}

\maketitle

\begin{abstract}

Active learning (AL) is a widely used technique for optimizing data labeling in machine learning by iteratively selecting, labeling, and training on the most informative data. However, its integration with formal privacy-preserving methods, particularly differential privacy (DP), remains largely underexplored. While some works have explored differentially private AL for specialized scenarios like online learning, the fundamental challenge of combining AL with DP in standard learning settings has remained unaddressed, severely limiting AL's applicability in privacy-sensitive domains. This work addresses this gap by introducing differentially private active learning (DP-AL) for standard learning settings. We demonstrate that naively integrating DP-SGD training into AL presents substantial challenges in privacy budget allocation and data utilization. To overcome these challenges, we propose \emph{step amplification}, which leverages individual sampling probabilities in batch creation to maximize data point participation in training steps, thus optimizing data utilization. Additionally, we investigate the effectiveness of various acquisition functions for data selection under privacy constraints, revealing that many commonly used functions become impractical. Our experiments on vision and natural language processing tasks show that DP-AL can improve performance for specific datasets and model architectures. However, our findings also highlight the limitations of AL in privacy-constrained environments, emphasizing the trade-offs between privacy, model accuracy, and data selection accuracy.\makeatletter{\renewcommand*{\@makefnmark}{}\footnotetext{This work has been accepted for publication in the \textit{IEEE Conference on Secure and Trustworthy Machine Learning (SaTML)}. The final version will be available on IEEE Xplore.}\makeatother}

\end{abstract}


\section{Introduction}
\label{sec:intro}

The development of accurate \emph{machine learning} (ML) models fundamentally relies on the availability of large labeled datasets. Although recent advances in self-supervised learning have reduced the need for labeled data in certain domains, labeled data remains essential for the vast majority of tasks.

While collecting raw data is often relatively straightforward, the process of labeling this data can be both expensive and time-consuming, particularly when expert knowledge is required. This constraint is especially pronounced in privacy-sensitive domains, where the cost of labeling is exceptionally high due to the need for skilled and trusted annotators. As a result, generating large and representative labeled datasets often becomes prohibitively expensive or even impossible. For instance, in the medical domain, although large volumes of patient data are collected, the high cost of expert annotation severely limits the availability of labeled data, thus reducing its utility for ML workflows. For example, annotating a single magnetic resonance imaging (MRI) scan can require several hours of work and consensus from multiple radiologists. Consequently, practitioners often resort to labeling \enquote{easy} subsets of data, which may not only be too small for training reliable and well-generalizing ML models, but also risk introducing a form of bias towards easy-to-label data. Datasets biased in this way potentially misrepresent the complexity of real-life tasks and the models trained on them risk underserving individuals whose data is hard to label.

\begin{figure}[t]
    \centering
    \includegraphics[width=.95\linewidth]{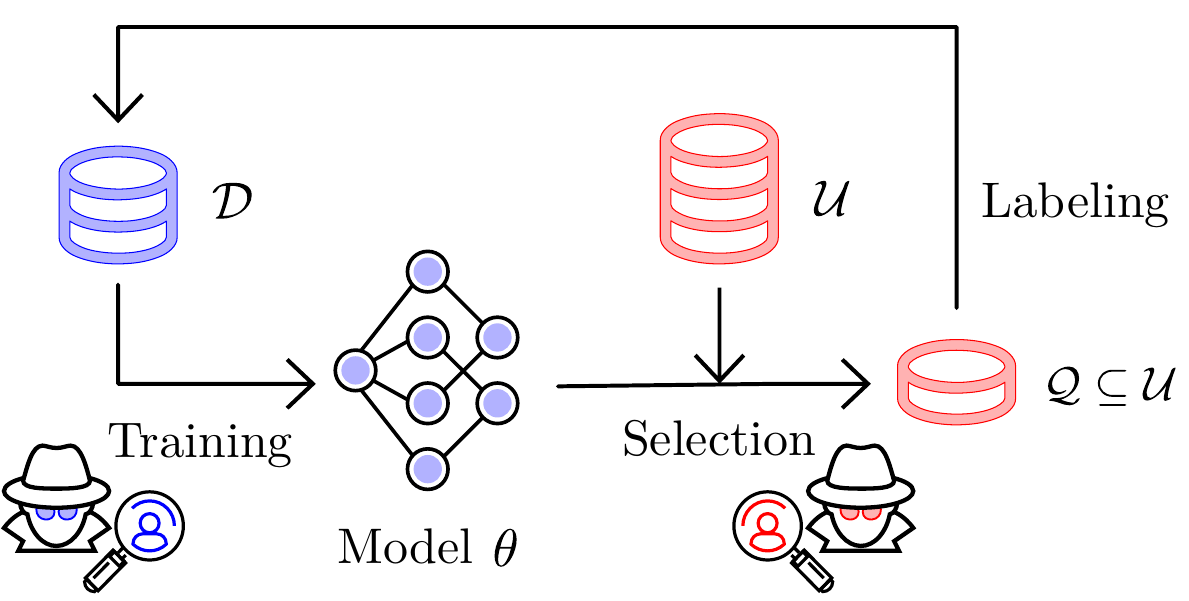}
    \caption{Overview of the iterative active learning process. First, the model is trained on the labeled dataset $\mathcal{D}$. Then, an acquisition function uses the current model to select the most informative samples from the unlabeled dataset ($\mathcal{Q} \subseteq \mathcal{U}$), which are labeled and added to the training dataset. The AL process exposes two privacy vulnerabilities: (1) as discussed in DP literature, an adversary may use the model and training gradients to infer private information, and (2) unique to AL, an adversary could exploit the results of the acquisition function to infer the presence of specific samples in the dataset.}
    \label{fig:overview}
\end{figure}

\emph{Active learning} (AL) has emerged as a promising solution to this problem by making the annotation process more efficient. AL enables models to achieve high performance with limited labeled data by iteratively selecting specific samples from an unlabeled dataset and requesting labels for only those instances. This strategy avoids labeling and training on samples that contribute little to improving the model.

However, when applied to sensitive domains like medicine, AL must adhere to strict privacy constraints \cite{Kaissis2020}. Privacy preservation is a significant concern in ML, as models can leak sensitive information about their training data, potentially compromising user privacy \cite{Geiping2020, Carlini2022}. This risk becomes even more pronounced when AL is applied, as the data selection process becomes a source for privacy leakage additional to what is typically addressed in conventional ML training (see \cref{fig:overview}). For example, many works have shown that ML models can be susceptible to membership inference attacks (MIAs), where an adversary attempts to determine whether a particular data point was part of the training dataset \cite{Shokri2017, Carlini2022}. In the context of AL, this vulnerability is exacerbated by the deterministic nature of the selection query. For an adversary with access to the unlabeled dataset and intermediate models (as is commonly assumed in strong threat models), inferring the membership of a data point in the training set is trivial: By \enquote{replaying} each selection phase by applying the model checkpoints to the unlabeled dataset, the adversary can identify which data points were most informative to that model checkpoint and thus added to the training dataset, resulting in a potential 100\% success rate for MIAs. This vulnerability to MIAs highlights a requirement to privatize not only the model training but also the data selection process.

To address privacy concerns in ML, \emph{differential privacy} (DP) \cite{Dwork2014} has become the gold standard privacy technique. DP guarantees that the statistical influence of an individual's data on an algorithm's output is bounded by a so-called \emph{privacy budget}. In ML training, methods like \emph{differentially private stochastic gradient descent} (DP-SGD) \cite{Abadi2016,Song2013} enforce these guarantees by adding noise to the gradients, preventing the model from learning identifying information about individuals. However, applying DP often results in reduced model utility, especially when working with limited data. This trade-off increases the importance of selecting and utilizing the most helpful data points during training, \eg, using AL.

The combination of AL and DP-SGD thus presents a promising approach to address both labeling costs and privacy concerns. However, several critical questions remain unanswered: How can we effectively integrate these two approaches? Will AL remain effective under privacy constraints? And: how might the inherent trade-offs in both techniques compound or mitigate each other?

In this work, we address these open questions by introducing a novel method for \emph{differentially private active learning} (DP-AL) that effectively combines AL with DP-SGD. Our research focuses on the classic pool-based AL setting (where a large \emph{pool} of unlabeled data is available) for classification tasks, though the principles can be extended to regression problems where \enquote{targets} replace \enquote{labels}. Our key contributions are:

\begin{enumerate}
    \item \textbf{Privacy-preserving Active Learning:}\: We propose a comprehensive method that enables active learning in privacy-constrained domains. \emph{Problems:} AL methods leak sensitive information through both the model training and the selection process. Existing private AL methods only consider weak privacy techniques or are limited to online learning scenarios with streaming data access. \emph{Solution:} We incorporate differential privacy into the training and selection phases of AL and jointly account their privacy losses to maximize their combined utility and privacy guarantees.
    \item \textbf{Step Amplification:} We propose a novel technique called \emph{step amplification} to optimize privacy budget utilization across the AL process. \emph{Problem:} Naive integration of DP and AL leads to inefficient use of the privacy budget and underutilization of informative data points. \emph{Solution:} Step amplification maximizes the number of training steps for each data point while ensuring equal privacy loss across all points by adapting their sampling probabilities in batch creation.
    \item \textbf{Critical Analysis of DP-AL Effectiveness:} We conduct a comprehensive evaluation to understand the impact of DP-AL. \emph{Problem:} The effectiveness and potential drawbacks of integrating privacy constraints with AL remain unclear. \emph{Solution:} We perform an in-depth analysis of DP-AL on diverse tasks, including image classification and natural language processing, under different privacy settings, highlighting the superiority of step amplification in most cases.
\end{enumerate}

An important outcome of our research is the finding that AL is not a universal solution to reducing labeling costs in privacy-constrained environments. While DP-AL can offer clear benefits in certain scenarios, it also introduces new challenges and trade-offs. Our findings thus underscore the complexity of balancing privacy, model accuracy, and effective data selection.

\section{Background and Related Work}\label{sec:related_work}

\subsection{Active Learning} 

Active learning (AL) is designed to improve model performance by iteratively selecting and labeling the most informative data points during training. Since it is often infeasible to label the entire collected dataset, a representative subset must be selected. Instead of selecting this subset randomly, AL aims to maximize the utility of the labeled data by selecting the data points that are expected to improve the model the most.

Let $\mathcal{D} = \{(x_i, y_i)\}_{i=1}^N$ denote the labeled dataset and $\mathcal{U}=\{x_j\}_{j=1}^M$ the unlabeled dataset. After an initial training phase using $\mathcal{D}$, the model evaluates the unlabeled dataset to select a set of $k$ data points $\mathcal{Q}\subseteq\mathcal{U}$ for labeling, based on an acquisition function $C$. In deep learning, this acquisition function can leverage the model's uncertainty or other criteria to find the most useful data points for the current model. The selection process can be formalized as:

\begin{equation}
	\mathcal{Q} = {\arg\max}_{\mathcal{S}\subseteq\mathcal{U}, |\mathcal{S}|= k}C(\mathcal{S},\theta),
\end{equation}
where $\theta$ represents the parameters of the current model.

Once the selected data points are labeled, they are added to the training dataset, and the model is trained on the extended dataset. This iterative process of selecting, labeling, and training continues until the labeling budget $B$ (\ie, the number of samples that can be labeled) is exhausted, incrementally refining the model with each iteration.

\subsection{Privacy-preserving Active Learning}

While AL has witnessed much interest by the general machine learning community, only few works have studied its combination with data-privacy technologies like differential privacy (DP). Several works integrate AL with federated learning (FL) \cite{Goetz2019,Ahmed2020,Deng2022, Kim2023, Zhou2023,Cao2023,Ahn2024} and homomorphic encryption \cite{Song2023,Kurniawan2022}. However, none of these approaches employ DP specifically. This is a crucial point, as federated learning alone does not ensure privacy \cite{Geiping2020, Boenisch2023} and homomorphic encryption is a security method that does not provide privacy guarantees to the training data. In contrast, our approach combines AL with DP, ensuring robust data privacy protection. We note that, although we focus on the centralized learning setting, our method can be seamlessly integrated into existing federated learning frameworks.

In the centralized learning setting, \cite{Ghassemi2016, Bittner2021} integrate DP into \emph{stream-based} AL (an online learning approach where data points are processed sequentially) for anomaly detection using a simple support vector machine (SVM) classifier. This approach employs the exponential mechanism \cite{Dwork2014} to privatize the AL selection query and DP-SGD to train the model. In contrast, our work focuses on the standard pool-based AL setting, which introduces several unique challenges and employs deep neural networks, rendering it more suitable for modern machine learning applications.

Feyisetan \etal \cite{Feyisetan2019a} address the issue of an untrusted human labeler in AL by applying local differential privacy (LDP) together with $k$-anonymity to privatize the data \emph{before} annotation. However, LDP tends to degrade performance for many data modalities, such as images, due to the substantial noise required to ensure privacy. Moreover, the issue of trusting labelers is not unique to AL but arises in all labeling settings. Furthermore, we contend that trust towards the labeler is of minor concern compared to privacy leakage through publication of trained models. This is because many labeling experts, such as physicians, are trusted professionals bound by confidentiality. We thus assume a trusted labeler throughout.

Finally, Zhao \etal \cite{Zhao2019} employ AL to select informative \emph{public} data to enhance the performance of DP models. Orthogonally to our work, their method does not require privatizing the AL process.

\subsection{Differential Privacy}\label{sec:dp}

Differential privacy (DP) \cite{Dwork2014} provides a mathematical framework to guarantee that the inclusion or exclusion of a single data point in/from a dataset does not significantly affect the output of an algorithm, thus not revealing identifying information about individuals.

\begin{definition}[($\eps,\delta$)-Differential Privacy]
    A randomized algorithm (mechanism) $\mech$ satisfies $(\varepsilon,\delta)$-DP if, for any pairs of adjacent datasets $\mathcal{D}\simeq\mathcal{D}'$ that differ in a single sample and all sets of outcomes $\mathcal{S}\subseteq \operatorname{Range}(\mech)$, it holds that:
    \begin{equation}
    {\mathrm{Pr}}[\mech(\mathcal{D}) \in \mathcal{S}]\leq e^\varepsilon {\mathrm{Pr}}[\mech(\mathcal{D}')\in\mathcal{S}] + \delta,
\end{equation}
where the pair ($\eps,\delta$) represents the privacy budget, controlling the level of privacy protection. When $\delta=0$, the mechanism is said to be (pure) $\varepsilon$-DP. 
For details on DP see Appendix \ref{apdx:dp}.
\end{definition}

\subsection{Improving DP-SGD Accuracy} 

Differentially private stochastic gradient descent (DP-SGD) \cite{Song2013,Abadi2016} has established itself as the main approach for privacy-preserving ML, extending the traditional stochastic gradient descent (SGD) algorithm. DP-SGD ensures DP by privatizing the training gradients before the model update. This is achieved by first clipping per-sample gradients to an upper norm bound $C$. Then, calibrated Gaussian noise $\mathcal{N}(0, \hat{\sigma}^2\matr{I})$ is added, where the noise multiplier $\hat{\sigma}=C\sigma$ depends on both the clipping threshold $C$ and the noise scale $\sigma$. The required noise scale $\sigma$ is determined by the privacy budget. A higher privacy budget allows for a lower noise scale, leading to more accurate gradient estimates and higher utility. Crucially, the post-processing theorem of DP (\cref{thm:post_processing}) ensures that any subsequent operations on the privatized gradients, including model updates, maintain the DP guarantee.

Given that training typically involves multiple iterations where the dataset is repeatedly accessed, each iteration must be privatized to ensure the cumulative privacy loss stays within the given privacy budget. Efficient privacy accounting and composition across iterations are therefore important areas of research, \eg, \cite{Mironov2017,Gopi2021}.

A fundamental component of DP-SGD is \emph{privacy amplification by subsampling}. By constructing random sub-samples instead of fixed mini-batches, \eg, through Poisson sampling, the privacy guarantees are amplified because every point has only a probabilistic chance of being used in training. 

Given a fixed (expected) batch size $b=p \cdot |\mathcal{D}|$, where $p$ is the likelihood of any data point being included in a sub-sample and $|\mathcal{D}|$ is the dataset size, larger datasets reduce the likelihood $p$, lowering the expected participation of individuals and thus their privacy loss at each iteration. This shows that a significant challenge arises when working with limited labeled data: Smaller datasets obtain lower privacy amplification, resulting in larger noise disruption or reduced training steps to stay within the privacy budget. In such scenarios, identifying useful data points for labeling is crucial to optimize each training step and avoid redundancy. While leveraging labeled \emph{public} data to augment the small private dataset is a promising approach \cite{Zhao2019,Gu2023}, it may not always be feasible due to potential distribution shifts. Moreover, the arbitrary use of \enquote{public} data for DP workflows has recently been scrutinized \cite{Tramer2024}. Thus, our work focuses on identifying and labeling the most informative data points from a larger set of unlabeled \emph{private} data under a constrained labeling budget.

Finally, despite DP-SGD's effectiveness in preserving privacy, the added noise perturbs the gradient estimates and the clipping introduces bias, which can degrade model accuracy \cite{Knolle2023}. To address this privacy-utility trade-off, researchers have explored various strategies, such as modifying model architectures \cite{Klause2022,Hoelzl2023} and large-scale training regimes \cite{Sander2023,Berrada2023,De2022}, which we incorporate into our work.

\section{Differentially Private Active Learning}

In this section, we first outline our threat model and the challenges of naively integrating DP-SGD into active learning (AL). We then propose step amplification as a solution and analyze privacy risks from AL selection queries across different acquisition functions. Finally, we jointly account for the privacy losses of DP-SGD and selection phases.

\subsection{Threat Model}

To ensure robust privacy protection, we adopt a worst-case threat model, aligned with standard practices in DP research. In the context of DP-AL, our threat model assumes a powerful adversary with full knowledge of the learning algorithm, including access to all privatized gradients and intermediate model states throughout training. The only information withheld from the adversary are the specific noise samples added to the gradients during the DP-SGD process. Furthermore, the adversary possesses complete information about the underlying dataset, with the exception of the presence of an individual data point, \ie, membership information. The primary objective of this adversary is to infer the membership status of a specific data point within the training dataset, leaking private information.

We also assume a trusted labeler who provides accurate annotations without compromising privacy. This assumption is not overly restrictive, as it aligns with many active learning scenarios, where labeling is costly and requires specialized expertise. In such scenarios, labeling is often performed by professionals who are bound by strict confidentiality agreements or professional ethics codes.

\subsection{Warm-up: A Naive DP-AL Implementation} \label{sec:naiveAL}

The integration of DP-SGD into the AL process presents several significant challenges. This section establishes a foundation for DP-AL by outlining these challenges and introducing our initial naive approach to addressing them. 

\paragraph{Base Method} In standard AL, models are typically trained to convergence in each training phase before selecting new data points for labeling. However, this approach cannot be used with DP-SGD due to the limited privacy budget, which is expended with each training step, resulting in a limited number of useful training steps. Thus, especially more complex models are not guaranteed to have converged by the time the privacy budget is exhausted. 
Therefore, the total number of available training steps must be split across training phases: In each training phase $i$, the model is trained for a predetermined number of $e$ epochs, resulting in $n_i = \lfloor (|\mathcal{D}_1|+(i-1)k)/b \rfloor$ training steps, where $|\mathcal{D}_1|$ is the initial dataset size, $k$ is the number of new samples added per selection phase, $b$ is the expected batch size, and $\lfloor\cdot\rfloor$ denotes the floor function. Since $n_i$ is strictly monotonically increasing, more steps are allocated to later phases when more data are available.

Another fundamental challenge of DP-AL is the dynamic training dataset size, which increases with each selection phase as newly labeled data points are added. This affects the sampling probabilities for batch creation and thus the parameters of the DP-SGD mechanism. Given a fixed (expected) batch size $b$, the sampling probability in phase $i$ is $q_i = b/(|\mathcal{D}_1|+(i-1)k)$. As the dataset grows, this probability decreases, resulting in better privacy guarantees in later training phases due to privacy amplification by subsampling (\cref{thm:priv_ampl}). The changing parameters of the DP-SGD mechanism puts our approach within the regime of \emph{heterogeneous composition}, which necessitates the use of advanced privacy accountants to accurately track and manage privacy expenditure. In our implementation, we employ Rényi Differential Privacy (RDP) \cite{Mironov2017}, although other state-of-the-art privacy accounting methods (such as numerical accounting \cite{Gopi2021}) could also be applied effectively.

An important observation in our approach is that data points not selected for labeling, and thus not used in training, do not incur any privacy loss from DP-SGD. The privacy leakage associated with the selection query is further discussed in \cref{sec:al_sel}.

Finally, the desired privacy budget and the corresponding level of data protection are typically determined by the data contributor before training. Thus, the target privacy budget ($\eps,\delta$) is predefined. Given these target parameters, along with the number of steps $n_i$ and the sampling probabilities $q_i$ for each training phase $i$, we compute the noise multiplier $\hat{\sigma}$ required to satisfy ($\eps,\delta$)-DP (converted from RDP) using a binary search algorithm, such as the one implemented in the \texttt{Opacus} library \cite{Yousefpour2021}. 

Our complete naive DP-AL algorithm, incorporating these initial considerations, is detailed in \cref{alg:al}. This algorithm serves as a foundation for our DP-AL framework and provides a starting point for more sophisticated approaches that we will explore in subsequent sections.

\begin{algorithm}[tb]
    \caption{\emph{Naive} active learning with DP-SGD. The function \texttt{get\_noiseMultiplier} adjusts the noise multiplier given target privacy parameters (see \cref{alg:getnoise}). \texttt{DPSGD} is a function that trains a model for a given number of steps using DP-SGD. To ensure end-to-end DP of the naive method, the selection procedure (step 7) must be privatized according to \cref{sec:al_sel,sec:combPLOSS}.}
    \label{alg:al}
	\begin{algorithmic}[1]
        \Require Target privacy parameters $(\varepsilon, \delta)$, initial model parameters $\theta_1$, initial labeled dataset $\mathcal{D}_1$, initial unlabeled dataset $\mathcal{U}_1$, epochs per training phase $e$, batch size $b$, query size $k$, labeling budget $B$
        \State $T \gets \lfloor B/k \rfloor$ \Comment{Num. selection phases}
        \State $q \gets [b/(|\mathcal{D}_1|+(i-1)k)]_{i=1}^{T+1}$ \Comment{Sample rates}
        \State $n \gets [e/q_i]_{i=1}^{T+1}$ \Comment{Training steps}
		\State $\hat{\sigma}\gets \texttt{get\_noiseMultiplier(}\varepsilon,\delta,n,q\texttt{)}$
		\For{$i = 1, \ldots, T$} 
        \State $\theta_{i+1} \gets \texttt{DPSGD(}\theta_{i}, \mathcal{D}_{i}, \hat{\sigma}, q_{i}, n_{i}\texttt{)}$ \Comment{Training}
		\State $\mathcal{Q}_i ={\arg\max}_{\mathcal{S}\subseteq\mathcal{U}_{i}, |\mathcal{S}|\leq k}C(\mathcal{S},\theta_{i+1})$ \Comment{Selection}
        \State $\mathcal{D}_{i+1} = \mathcal{D}_{i} \cup \mathcal{Q}_i$
        \State $\mathcal{U}_{i+1} = \mathcal{U}_{i} \setminus \mathcal{Q}_i$
		\EndFor
        \State $\theta_{\text{final}} \gets \texttt{DP-SGD(}\theta_{T+1}, \mathcal{D}_{T+1}, \hat{\sigma}, q_{T+1}, n_{T+1}\texttt{)}$
	\end{algorithmic}
\end{algorithm}

\begin{figure*}[tb]
    \centering
    \begin{tikzpicture}
        \begin{axis}[
        axis x line=center,
        axis y line=left,
        xmin=1,
        xmax=5,
        ymin=0, 
        ymax=8,
        clip=true,
        xlabel={Training phase $i$},
        ylabel={$\varepsilon_i$},
        legend columns=-1,
        legend style={
            draw=none, fill=none, 
            font=\small,
            at={(1.05,1.35)}
        },
        width=0.48\linewidth,height=4.cm,
        domain=1:5]
          \addplot [Pred, mark=*, mark size=1.5pt] table [x=iter, y=g0] {plot_data/al_eps.txt};
          \addplot [Pgreen, mark=*, mark size=1.5pt] table [x=iter, y=g1] {plot_data/al_eps.txt};
          \addplot [Porange, mark=*, mark size=1.5pt] table [x=iter, y=g2] {plot_data/al_eps.txt};
          \addplot [Plila, mark=*, mark size=1.5pt] table [x=iter, y=g3] {plot_data/al_eps.txt};
          \addplot [QPblue, mark=*, mark size=1.5pt] table [x=iter, y=g4] {plot_data/al_eps.txt};
          \addplot [black, thick, dashed,opacity=0.7] {0*x+8};
          \legend{$\kset_1$, $\kset_2$, $\kset_3$, $\kset_4$, $\kset_5$, Budget}
        \end{axis}
    \end{tikzpicture}
    \hfill
    \begin{tikzpicture}
        \begin{axis}[
        axis x line=center,
        axis y line=left,
        xmin=1,
        xmax=5,
        ymin=0, 
        ymax=8,
        clip=true,
        xlabel={Training phase $i$},
        ylabel={$\varepsilon_i$},
        legend columns=-1,
        legend style={
            draw=none, fill=none, 
            font=\small,
            at={(1.05,1.35)}
        },
        width=0.48\linewidth,height=4.cm,
        domain=1:5]
          \addplot [Pred, mark=*, mark size=1.5pt] table [x=iter, y=g0] {plot_data/stepAmpl_eps.txt};
          \addplot [Pgreen, mark=*, mark size=1.5pt] table [x=iter, y=g1] {plot_data/stepAmpl_eps.txt};
          \addplot [Porange, mark=*, mark size=1.5pt] table [x=iter, y=g2] {plot_data/stepAmpl_eps.txt};
          \addplot [Plila, mark=*, mark size=1.5pt] table [x=iter, y=g3] {plot_data/stepAmpl_eps.txt};
          \addplot [QPblue, mark=*, mark size=1.5pt] table [x=iter, y=g4] {plot_data/stepAmpl_eps.txt};
          \addplot [black, thick, dashed,opacity=0.7] {0*x+8};
          \legend{$\kset_1$, $\kset_2$, $\kset_3$, $\kset_4$, $\kset_5$, Budget}
        \end{axis}
    \end{tikzpicture}
    \caption{Privacy loss across training phases for the naive (\emph{left}) and step amplification DP-AL method (\emph{right}) with a total privacy budget of $\varepsilon=8$. $\kset_i$ denotes the group of samples added to the training dataset prior to phase $i$. The figure shows that each training phase expends a different amount of privacy due to the change in sampling probabilities. In step amplification, contrary to the naive approach, all data points consume their full privacy budget.}
    \label{fig:eps_impl}
\end{figure*}
\begin{figure*}[tb]
    \centering
    \begin{tikzpicture}
        \begin{axis}[
        axis x line=center,
        axis y line=left,
        xmin=1,
        xmax=5,
        ymin=0, 
        ymax=.5,
        clip=true,
        xlabel={Training phase $i$},
        ylabel={Sampling prob. $q$},
        legend columns=-1,
        legend style={
            draw=none, fill=none, 
            font=\small,
            at={(1.05,1.35)}
        },
        width=.48\linewidth,height=4.cm,
        domain=0:4]
          \addplot [Pred, mark=*, mark size=1.5pt] table [x=iter, y=naive] {plot_data/sample_probs.txt};
          \addplot [QPblue, mark=*, mark size=1.5pt] table [x=iter, y=ourG1] {plot_data/sample_probs.txt};
          \addplot [Pblue, mark=*, mark size=1.5pt] table [x=iter, y=ourG2] {plot_data/sample_probs.txt};
          \legend{Naive, SA $\gset_{\text{old}}$, SA $\gset_{\text{new}}$}
        \end{axis}
    \end{tikzpicture}
    \hfill
    \begin{tikzpicture}
        \begin{axis}[
        axis x line=center,
        axis y line=left,
        xmin=1,
        xmax=5,
        ymin=0, 
        ymax=350,
        clip=true,
        xlabel={Training phase $i$},
        ylabel={Steps $n$},
        legend columns=-1,
        legend style={
            draw=none, fill=none, 
            font=\small,
            at={(1.05,1.35)}
        },
        width=.48\linewidth,height=4.cm,
        domain=0:4]
          \addplot [Pred, mark=triangle*] table [x=iter, y=naive] {plot_data/steps.txt};
          \addplot [QPblue, mark=*, mark size=1.5pt] table [x=iter, y=stepAmpl] {plot_data/steps.txt};
          \legend{Naive, SA}
        \end{axis}
    \end{tikzpicture}
    \caption{Sampling probabilities (\emph{left}) and number of training steps (\emph{right}) across training phases for the naive and step amplification (SA) DP-AL method. $\gset_{\text{old}}$ denotes the group of samples in the labeled dataset at phase $i-1$ and $\gset_{\text{new}}$ the new samples added at phase $i$. }
    \label{fig:steps_probs}
\end{figure*}

\paragraph{Inefficient Privacy Budget Utilization} The naive approach to DP-AL suffers from inefficient use of the privacy budget. Specifically, data points added to the training dataset in later phases experience substantially less privacy leakage compared to the initial data. This inefficiency is two-fold: First, samples added in later stages participate in fewer training steps, leading to less cumulative privacy leakage (\cref{fig:eps_impl} left). Second, as the training dataset grows, the sampling probability for these later samples decreases, meaning they benefit more from subsampling privacy amplification (\cref{fig:steps_probs} left).

To illustrate, let $\eps$ denote the total privacy budget and $[\varepsilon_1, \varepsilon_2, \ldots, \varepsilon_{T+1}]$ represent the sequence of privacy budgets allocated across the $T+1$ training phases. The initially labeled data points, used in every phase, accumulate a privacy loss equal to the total privacy budget: $\sum_{i=1}^{T+1} \varepsilon_i = \eps$. In stark contrast, the data points added in the final phase accumulate only $\varepsilon_{T+1}$, which is significantly smaller than $\eps$.\footnote{We overload $\eps$ throughout to also denote the RDP parameter, and disregard the RDP order for simplicity.}

However, the training process terminates when the initial samples exhaust their privacy budget, leaving many data points underutilized. \Cref{fig:eps_impl} (left) illustrates the distribution of privacy budget across different training phases and data points in a realistic scenario. The graph clearly highlights that the privacy budget is primarily consumed by the initial samples, while data points added in later phases incur minimal privacy loss. This leads not only to an imbalance in privacy loss but also to reduced model performance, as the underutilization of later-added data points limits the model's ability to fully learn from important data.

\subsection{Step Amplification: Optimally Utilising the Privacy Budget} \label{sec:stepAmp}

\begin{algorithm}[tb]
     \caption{Step amplification algorithm. $n_{:i}$ denotes a sequence containing the number of training steps up to phase $i$. $\gset_{\text{old,i}}$ and $\gset_{\text{new,i}}$ represent the set of data points from the previous training dataset ($\dset_{i-1}$) and the newly labeled data points ($\mathcal{Q}_{i-1}$), respectively. $\lfloor \cdot \rceil$ denotes rounding to nearest integer. \texttt{get\_sampleRate} is a function that computes the sampling rate corresponding to the given privacy parameters (see \cref{alg:getsample}). \texttt{get\_epsilon} is a function that computes the privacy loss from DP-SGD training (see \cref{alg:geteps}).}
 \label{alg:stepampl}
	\begin{algorithmic}[1]
        \Require Target privacy parameters $(\varepsilon, \delta)$, noise multiplier $\sigma$, number of AL iterations $T$, expected batch size $b$, set of sample rates for AL iterations $q$, set of training steps for AL iterations $n$
        \For{$i = 2, \ldots, T+1$}
        \State $\varepsilon_{\text{target}} \gets \texttt{get\_epsilon(}\delta, \sigma, n_{:i}, q_{:i}\texttt{)}$
        \State $q_{\text{old},i} \gets \texttt{get\_sampleRate(}\varepsilon_{\text{target}},\delta,\sigma,n_{:i}, q_{:{i-1}}\texttt{)}$
        \State $q_{\text{new},i} \gets \texttt{get\_sampleRate}(\eps_{\text{target}},\delta,\sigma,n_{i})$
        \State \textbf{init} $n_{i,\text{low}}\gets n_i$, $n_{i,\text{high}}\gets 3n_i$
        \While{$b \not\approx q_{\text{old},i}|\gset_{\text{old},i}|+q_{\text{new},i}|\gset_{\text{new},i}|$}
        \State $n_i \gets \lfloor (n_{i,\text{high}} - n_{i,\text{low}})/2 \rceil$
        \State $q_{\text{old},i} \gets \texttt{get\_sampleRate(}\varepsilon_{\text{target}},\delta,\sigma,n_{:i},q_{:{i-1}}\texttt{)}$
        \State $q_{\text{new},i} \gets \texttt{get\_sampleRate}(\varepsilon_{\text{target}},\delta,\sigma,n_{i})$
        \If{$q_{\text{old},i}|\gset_{\text{old},i}|+q_{\text{new},i}|\gset_{\text{new},i}| > b$}
            \State $n_{i,\text{low}} \gets n_i$
        \Else
            \State $n_{i,\text{high}} \gets n_i$
        \EndIf
        \EndWhile
        \EndFor
    \State \Return $n, q_{\text{old}}, q_{\text{new}}$
	\end{algorithmic}
\end{algorithm}

To address the poor use of data and privacy budget in the naive approach, we propose a novel method called \emph{step amplification}. Step amplification optimizes data utilization by ensuring that \textbf{all data points incur an equal privacy loss by the end of each training phase}. This is achieved by adjusting the sampling rate for each group of data points to maximize the expected number of update steps they participate in.

Consider the sequence of privacy budgets $[\varepsilon_1, \varepsilon_2, \ldots, \varepsilon_{T+1}]$ allocated across training phases. In each phase $i$, the current dataset $\mathcal{D}_i = \mathcal{D}_{i-1}\cup \mathcal{Q}_{i-1}$ can be split into two groups: $\gset_{\text{old}}$, which includes the previous labeled dataset $\mathcal{D}_{i-1}$, and $\gset_{\text{new}}$, which consists of the $k$ newly annotated data points in $\mathcal{Q}_{i-1}$. The points in $\gset_{\text{old}}$ are restricted to the phase's allocated privacy budget $\eps_{\text{old}}\gets \eps_i$ due to privacy leakage in prior phases. In contrast, $\gset_{\text{new}}$ has not been employed in training yet and can also use the accumulated budget from prior phases, \ie, $\eps_{\text{new}} \gets \sum_{j=1}^i \eps_j$. Despite this initial disparity where $\eps_{\text{old}} < \eps_{\text{new}}$, the methodology ensures that overall privacy loss equalizes to $(\sum_{j=1}^i \eps_j)$ for all data points after training phase $i$, as visualized in \cref{fig:eps_impl} (right).

Managing different privacy budgets between the data groups presents two key challenges: First, methodological changes to the training process are required, as simple approaches such as exclusive training on new data could lead to catastrophic forgetting (\cref{apdx:dispLeakage}). To address this, we modify the batch creation method of DP-SGD by introducing a sampling schedule. Rather than sampling the points in each batch with uniform probability $q$, we assign a higher sampling probability to the new data compared to the old data ($q_{\text{new}} > q_{\text{old}}$). This technique, which is similar to individual privacy assignment \cite{Boenisch2023b}, effectively increases the utilization of new data by overrepresenting them in the batches and allowing them to \enquote{catch up} with the privacy loss of the other points. 

However, this approach increases the expected batch size $b = q_{\text{old}}|\gset_{\text{old}}| + q_{\text{new}}|\gset_{\text{new}}|$, which is undesirable (\eg, due to hardware constraints). Thus, the second challenge involves maintaining the original expected batch size while adjusting the sampling probabilities. To address this, we must modify additional DP-SGD parameters, specifically either the noise multiplier $\hat{\sigma}$ or the number of training steps $n_i$. \Eg, increasing $n_i$ results in a larger privacy loss, which can be offset by reducing sampling probabilities for both groups to satisfy the privacy budget and decrease the expected batch size.

We experimentally find that adjusting the number of training steps yields superior results (\cref{apdx:dispLeakage}). However, since $n_i$ must be a natural number, step modifications alone might not achieve the target expected batch size---particularly when adjusting $n_i$ by $\pm 1$ leads to sampling probabilities that cause substantial deviations from the target in either direction. In such cases, we make minor adjustments to the noise multiplier during this training phase. \Cref{fig:steps_probs} illustrates the evolution of sampling probabilities and training steps under this approach.

Intuitively, we implement step amplification using the following steps (also see \cref{alg:stepampl}): (1) For each training phase $i>1$, we re-compute the cumulative target privacy budget $\eps_\text{target} \gets \sum_{j=1}^i\eps_j$ by composing DP-SGD mechanisms from the current and prior phases. This prevents the accumulation of approximation errors across phases. (2) We determine sampling probabilities for the old and newly labeled data groups that satisfy their individual privacy budgets. Rather than explicitly computing $\eps_\text{new}$ and $\eps_\text{old}$, we employ composition of step and sampling probability schedules---applying the full history of DP-SGD parameters to the old data group while restricting the new data group to current phase parameters. (3) When sampling probabilities yield an expected batch size that deviates substantially from the target, we employ binary search to iteratively optimize the number of training steps $n_i$ until achieving sampling probabilities that produce an acceptable expected batch size. For clarity, we omit adjustments to the noise multiplier in \cref{alg:stepampl}. (4) The results of step amplification are: increased number of training steps per phase and adjusted sampling probabilities for old and new data points. Other DP-SGD parameters remain unchanged.

In summary, step amplification is as a powerful improvement to the naive approach, substantially enhancing privacy budget utilization. Its computational efficiency stems from front-loading parameter calculations, allowing subsequent training to proceed without additional overhead (see \cref{apdx:runtime}). We integrate step amplification seamlessly into the naive framework by first executing steps 1-4 from the naive algorithm (\cref{alg:al}), then applying step amplification, before proceeding with step 5. The only change to the training process involves batch creation with varying sampling probabilities as described above.

Finally, we note that step amplification is data-independent, thus, no additional privacy leakage is incurred. 

\subsection{Privatizing the Selection Phases} \label{sec:al_sel}

Having addressed the training stage of the AL workflow, we now turn to the data selection stage of AL. Our aim is to investigate how the AL selection can be privatized while maintaining its effectiveness.

To this end, we analyzed different types of acquisition functions and discovered that many methods which are successful for non-private AL are unsuitable for DP-AL. In the following, we present our findings, starting with uncertainty sampling (the most promising approach) and then discuss two methods we found unsuitable: diversity sampling and training multiple models.

\paragraph{Uncertainty Sampling}

Uncertainty sampling is a common strategy in AL that includes various acquisition functions. This approach involves selecting data points for labeling that the current model $f_{\theta}$ is least certain about. The uncertainty is derived from the model's output logits $f_{\theta}(\vect{x})$. The selection process can be formulated as a top-$k$ query:

\begin{equation}
	\mathcal{Q} = {\arg\max}_{\mathcal{S}\subseteq\mathcal{U}, |\mathcal{S}| = k} \sum_{\vect{x}\in \mathcal{S}} C(f_{\theta}(\vect{x})),
\end{equation}
where $C$ represents the specific acquisition function. Note that more than one subset may have the same summed score, in which case ties can be broken, \eg, by random selection. 

In uncertainty sampling, each score $C(f_{\theta}(\vect{x}))$ is computed independently for each data point in the unlabeled pool $\mathcal{U}$. This independence allows the decomposition of the selection query into parallel sub-queries, which are then privatized individually based on \emph{parallel composition} (\cref{thm:parallel_comp}). The subsequent aggregation using the maximum operation falls under post-processing, which (according to the post-processing guarantee of DP in \cref{thm:post_processing}) does not introduce additional privacy loss. Importantly, because the selection query uses the privatized model and the unlabeled data without re-using any training data, the privacy guarantees of the training data remains unchanged.

Thus, the only quantities which require privatization are the uncertainty scores. For this, we use the \emph{Laplace mechanism} which is well-suited due to its effectiveness in low-dimensional queries. We empirically confirm this in Appendix \ref{apdx:sel_mechanisms}, where we test multiple mechanisms for privatization and find the Laplace mechanism to perform best.

To further improve practical utility, we leverage a \enquote{clipping trick}. Neural networks tend to be overconfident in their predictions, leading to generally low uncertainty estimates. As the noise added for DP is calibrated to a function's global sensitivity, it can disproportionately affect these lower values. By clipping the acquisition function's outputs to some data-independent value, we reduce its range without losing significant signal (as high values are practically never observed). Intuitively, this approach increases the signal-to-noise-ratio of the privatized uncertainties. We provide a detailed analysis of the global sensitivities of specific uncertainty-based acquisition functions in Appendix \ref{apdx:acFunctions}.

\paragraph{Unsuitable Acquisition Functions}

We now discuss two commonly used techniques in non-private AL which turn out to be unsuitable under DP.

\emph{Diversity sampling} in AL selects data points that maximize the diversity of the labeled dataset, ensuring a wide range of representative samples. Diversity approaches typically employ iterative clustering methods, resulting in different privacy leakage characteristics compared to uncertainty sampling approaches. As an example, we consider two prominent methods from this category: CoreSet \cite{Sener2018} and BADGE \cite{Ash2020}.

CoreSet utilizes a greedy $k$-center algorithm, which sequentially selects data points farthest from the labeled data in the embedding space of the current model. BADGE computes gradients for unlabeled data with respect to the predicted class and clusters these gradient vectors using $k$-means++ initialization, also employing a greedy sequential selection method. Both methods rely on sequential selection and informativeness measures (distances) that depend on all unlabeled data points (in CoreSet, even on already labeled points), making them unsuitable for the parallel composition approach used in uncertainty sampling.

To illustrate the challenges of implementing diversity methods under DP constraints, we present a simplified example based on CoreSet: Consider a scenario with $|\mathcal{U}| = 1000$ unlabeled data points, a labeling budget of $k = 100$, and a privacy budget for selection $\epssel=1$. As the methods require $k$ sequential selections, we must divide the privacy budget equally across selection phases $j$ (by \cref{thm:basic_comp}), resulting in $\epssel(j) = \epssel/k = 0.01$ per selection. Using, for example, the exponential mechanism to privatize each selection, we have:

\begin{equation}
\prob(j) \propto \exp\left(\frac{\epssel(j) d_{\text{min}}(j)}{2\Delta}\right),
\end{equation}
where $d_{\text{min}}(j)$ is the minimum distance to the labeled set, and $\sens{}$ is the global sensitivity of the utility function. Assuming, without loss of generality, that $\sens{}=1$ (normalized distance in embedding space), a single selection is given by:

\begin{equation}
\prob(j) \propto \exp(0.005 \cdot d_{\text{min}}(j)).
\end{equation}

For three points with distances of 0.1, 0.5, and 1, this results in selection probabilities of approximately 0.3325, 0.3333, and 0.3342, respectively. These near-uniform probabilities effectively reduce the diversity selection to random sampling.

Therefore, we conclude that diversity-based approaches become impractical under DP constraints due to the significant privacy budget requirements and their inherent sequential nature. The privacy budget dilution across multiple selections effectively negates the intended benefits of AL, reducing them to approximate random sampling.

Another prominent approach for AL involves \emph{training additional models} alongside the target model. For instance, query-by-committee (QBC)\cite{Seung1992} trains multiple models on the same dataset and the disagreement between models is used as a measure of uncertainty. Another approach, GCNAL \cite{Caramalau2021}, trains an additional graph convolutional network to measure the relation between labeled and unlabeled instances. Unfortunately, these methods are not applicable in the context of DP-AL, as the training of additional models would also need to be privatized. This necessitates splitting the privacy budget across multiple trainings, which can significantly compromise the accuracy of each individual model.

Overall, we find that several selection methods used in non-private AL interact poorly with DP, making more advanced acquisition methods like diversity sampling and training multiple models that do not allow for parallel composition non-viable. We find uncertainty sampling, which allows for parallel privatization of the selection queries, to be the most suitable approach for DP-AL. From here on, we focus our investigation on privatizing uncertainty-based acquisition functions.

\subsection{Joint Privacy Accounting of DP-SGD and Selection} \label{sec:combPLOSS}

Until now, we have considered the privacy leakage from the DP-SGD training phases and the AL selection phases independently. However, to determine the overall privacy loss in the DP-AL process, we must \emph{aggregate} the privacy leakage from all phases.

As previously discussed, different groups of data points incur varying degrees of privacy loss depending on when they were labeled and which phases they participated in. For instance, consider a group of samples labeled during the $j$-th selection phase. These samples would participate in $j$ out of the $T$ selection phases and subsequently participate in the remaining $T+1-j$ training phases (since there are $T+1$ training phases in total).

To protect privacy during the selection phases, we utilize the Laplace mechanism alongside basic composition (\cref{thm:basic_comp}). Assuming a privacy budget for selection of $\eps_{\text{Sel}}$, the privacy loss incurred during each phase is $\eps_{\text{Sel}}/T$. Thus, a group of samples added during the $j$-th selection phase accumulates a privacy loss of $j\eps_{\text{Sel}}/T$ before they even participate in training. This has to be accounted for in our step amplification method. Importantly, data points which are never selected incur a total privacy loss of $\eps_{\text{Sel}}$, with no privacy loss from training.

With step amplification we ensure that all data points used in training accumulate the same privacy loss after completing each training phase (see \cref{fig:eps_impl}). Formally, given the sequence of privacy budgets $[\varepsilon_1, \varepsilon_2, \ldots, \varepsilon_{T+1}]$ across training phases, the group of points already used in $i$ training phases have accumulated a privacy loss of $\sum_{l=1}^{i}\eps_l$ and can only use the budget allocated for the next phase, \ie, $\eps_{\text{old}}=\eps_{i+1}$. Conversely, newly added data points, which have not yet incurred any training-related privacy loss, can additionally utilize the accumulated privacy budget from all previous phases, $\eps_{\text{new}}=\sum_{l=1}^{i+1}\eps_l$.

However, when new data points enter training with an existing privacy loss of $j\epssel/T$ from the $j$ selection phases, we adjust their privacy budgets for the next training phase using basic composition. Specifically, their budget is reduced as follows:

\begin{equation}
\eps'_{\text{new}} = \eps_{\text{new}}-j\frac{\eps_{\text{Sel}}}{T}.
\end{equation} 

A visualization of how privacy loss accumulates across training phases in the presence of selection privacy loss is shown in \cref{fig:eps_impl2}.

In cases where the acquisition function incorporates training data points into the selection process, all data points are involved in every selection phase. Thus, each data point accumulates a privacy loss of $\eps_{\text{Sel}}$ from selection alone. Consequently, the DP-SGD privacy budget must be reduced to $\eps - \eps_{\text{Sel}}$, rather than simply aggregating partial losses as previously described. The greater impact on privacy of such acquisition functions makes them less favorable for DP-AL.

Our proposed DP-AL method, incorporating step amplification and joint privacy accounting of training and selection phases, is presented in \cref{alg:overall}.

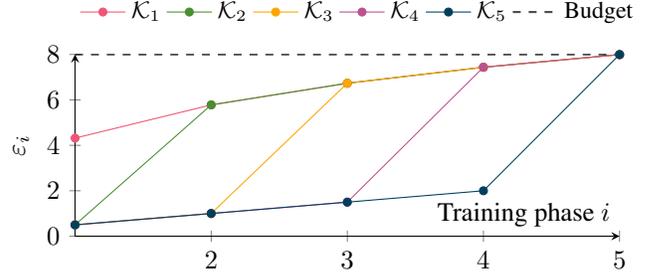
\begin{figure}[tb]
    \centering
    \begin{tikzpicture}
        \begin{axis}[
        axis x line=center,
        axis y line=left,
        xmin=1,
        xmax=5,
        ymin=0, 
        ymax=8,
        clip=true,
        xlabel={Training phase $i$},
        ylabel={$\varepsilon_i$},
        legend columns=-1,
        legend style={
            draw=none, fill=none, 
            font=\small,
            at={(1.05,1.35)}
        },
        width=\linewidth,height=4.cm,
        domain=1:5]
          \addplot [Pred, mark=*, mark size=1.5pt] table [x=iter, y=g0] {plot_data/stepAmplDP_eps.txt};
          \addplot [Pgreen, mark=*, mark size=1.5pt] table [x=iter, y=g1] {plot_data/stepAmplDP_eps.txt};
          \addplot [Porange, mark=*, mark size=1.5pt] table [x=iter, y=g2] {plot_data/stepAmplDP_eps.txt};
          \addplot [Plila, mark=*, mark size=1.5pt] table [x=iter, y=g3] {plot_data/stepAmplDP_eps.txt};
          \addplot [QPblue, mark=*, mark size=1.5pt] table [x=iter, y=g4] {plot_data/stepAmplDP_eps.txt};
          \addplot [black, thick, dashed,opacity=0.7] {0*x+8};
          \legend{$\kset_1$, $\kset_2$, $\kset_3$, $\kset_4$, $\kset_5$, Budget}
        \end{axis}
    \end{tikzpicture}
    \caption{Privacy loss across training phases for our DP-AL implementation with a total privacy budget of $\varepsilon=8$ and a privacy budget from selection of $\epssel=2$. $\kset_i$ denotes the group of samples added to the training dataset prior to phase $i$. Newly added samples already use some of the privacy budget before training, due to the privacy leakage from the selection phases they where used in.}
    \label{fig:eps_impl2}
\end{figure}

\section{Experiments}\label{sec:experiments}

In this section, we investigate the effectiveness of active learning (AL) with differentially private stochastic gradient descent (DP-SGD), including our proposed step amplification method. For this, we employ various image classification and natural language processing (NLP) tasks. The source code is publicly available at \url{https://github.com/kschwethelm/DPAL}.

\subsection{Experimental Setting}

\begin{table*}[t]
    \centering
    \caption{Comparison of AL methods under differential privacy (DP): Performance evaluation of various AL approaches with a privacy budget of $\eps=8$. \emph{Random} denotes standard DP-SGD training, \emph{Random AL Naive} represents naive AL training, and \emph{Random AL} employs our step amplification (SA) method, as do the remaining methods. Note that non-DP methods use DP-SGD but do not privatize the AL selection query. They are included to assess the upper bound utility of DP-AL. Area Under Curve (AUC) for CheXpert multi-label classification was computed per class and then averaged over classes. Results are reported as mean $\pm$ standard deviation over 5 training runs, except for CheXpert, which was run 3 times due to computational constraints. The best performances of DP methods are highlighted in bold and the highest scores overall are underlined.}
    \label{tab:performance}
    \begin{tabular}{lcc|cc|ccc|c}
        \toprule
          & & & CIFAR-10 & CIFAR-10 & BloodCell & RetinalOCT  & CheXpert & SNLI \\
         & & & ResNet-9 & Eq-ResNet-9 & ResNet-9 & ResNet-9 & NFNet-F0 & BERT \\\midrule
         Method & SA & DP & Accuracy (\%) & Accuracy (\%) & Accuracy (\%) & Accuracy (\%) & AUC (\%) & Accuracy (\%)\\\midrule
         Random & $-$ & $\checkmark$ & $66.58_{\pm 0.94}$ & \underline{\bm{$76.45_{\pm 0.62}$}} & \underline{\bm{$92.31_{\pm 0.94}$}} &  $74.54_{\pm 3.37}$  & $85.74_{\pm 0.71}$ & $78.62_{\pm 0.16}$ \\
         Random Naive AL & $\times$ & $\checkmark$ & $63.40_{\pm 0.67}$ & $72.87_{\pm 0.54}$ & $87.02_{\pm 1.94}$ & $69.84_{\pm 2.47}$ &  $84.43_{\pm 0.64}$ & $78.11_{\pm 0.20}$ \\
         Random AL & $\checkmark$ & $\checkmark$ & $66.55_{\pm 1.22}$ & $73.23_{\pm 0.60}$ & $88.52_{\pm 0.90}$ & $73.50_{\pm 4.28}$ &  $85.59_{\pm 0.89}$ & $78.66_{\pm 0.34}$ \\\midrule
         Least Confidence & $\checkmark$ & $\times$ & $68.32_{\pm 0.66}$ & $74.87_{\pm 0.49}$  & $90.81_{\pm 1.36}$ & $76.58_{\pm 3.65}$ &  $86.83_{\pm 0.33}$ & $79.21_{\pm 0.15}$ \\
         Minimum Margin & $\checkmark$ & $\times$ & $68.19_{\pm 0.48}$ & $74.60_{\pm 0.45}$ & $91.82_{\pm 1.02}$ & $80.73_{\pm 3.57}$ & $-$ & \underline{$79.40_{\pm 0.12}$} \\
         Entropy & $\checkmark$ & $\times$ & \underline{$68.43_{\pm 0.59}$} & $74.99_{\pm 0.47}$ & $91.22_{\pm 1.21}$ & \underline{$81.48_{\pm 2.79}$} & \underline{$87.11_{\pm 0.23}$} & $79.37_{\pm 0.19}$ \\
         MC BALD & $\checkmark$ & $\times$ & $67.09_{\pm 0.94}$ & $74.01_{\pm 0.61}$ & $85.10_{\pm 0.91}$ & $76.95_{\pm 2.98}$ & $85.94_{\pm 0.55}$ & $78.64_{\pm 0.19}$ \\\midrule
         Least Confidence & $\checkmark$ & $\checkmark$ & $66.91_{\pm 0.63}$ & $73.41_{\pm 0.62}$ & $88.85_{\pm 0.74}$ & $75.92_{\pm 3.11}$ & $86.49_{\pm 0.37}$ & $78.83_{\pm 0.30}$ \\
         Minimum Margin & $\checkmark$ & $\checkmark$ & $66.40_{\pm 0.50}$ & $73.45_{\pm 0.62}$ & $89.44_{\pm 0.63}$ & $77.60_{\pm 3.34}$ & $-$ & \bm{$78.90_{\pm 0.27}$} \\
         Entropy & $\checkmark$ & $\checkmark$ & \bm{$67.30_{\pm 0.82}$} & $73.64_{\pm 0.33}$ & $89.07_{\pm 1.05}$ & \bm{$78.00_{\pm 3.34}$} & \bm{$86.71_{\pm 0.42}$} & $78.86_{\pm 0.16}$ \\
         MC BALD & $\checkmark$ & $\checkmark$ & $66.12_{\pm 0.94}$ & $72.98_{\pm 0.54}$ & $87.99_{\pm 0.82}$ & $76.37_{\pm 3.14}$ & $86.08_{\pm 0.60}$ & $78.22_{\pm 0.14}$ \\
         \bottomrule
    \end{tabular}
\end{table*}

This section outlines our experimental setting, with further details provided in Appendix \ref{apdx:exp_details}. 

\paragraph{Methods} We evaluate the performance of several state-of-the-art AL acquisition functions and compare them against random sampling as a baseline for labeling data points. The uncertainty-based methods considered include least confidence \cite{Culotta2005}, minimum margin \cite{Scheffer2001}, entropy \cite{Joshi2009}, and Bayesian active learning by disagreement (BALD) with Monte Carlo dropout \cite{Houlsby2011,Gal2017}. See Appendix \ref{apdx:acFunctions} for an overview of these acquisition functions. 

For comparison, we establish several baselines based on random sampling. Our primary baseline is standard DP-SGD training without AL, where the training dataset is randomly sampled from the unlabeled dataset \emph{before} training. This baseline represents the standard approach when labeling the entire dataset is infeasible. If no AL method can outperform this baseline, there is no justification for using AL in privacy-constrained settings. To investigate the impact of the incremental training and iterative data selection process inherent to AL, we include a variant that employs our DP-AL framework (including step amplification) but \emph{selects random data points} during the selection phases. This setup additionally highlights the contribution of the acquisition functions. Finally, we demonstrate the efficacy of step amplification (\cref{sec:stepAmp}) vs. our naive DP-AL approach (\cref{sec:naiveAL}), both using random sampling. Notably, none of the aforementioned random sampling methods incur privacy leakage from selection and can thus utilize the entire privacy budget for training.

To further understand the effects of privatizing the selection phases, we also conduct experiments under the assumption that the acquisition functions do not pose a privacy risk, allowing us to maximize their accuracy and use the full privacy budget for DP-SGD alone. It is important to note that these methods do not satisfy DP, they are only included to illustrate the upper bounds of performance in the absence of selection privacy constraints.

\paragraph{Datasets} We simulate the AL process by using already labeled datasets while treating the labels as initially unknown and imposing a labeling budget smaller than the full dataset size. After each AL selection round, the labels of the chosen points are revealed. 

We evaluate general performance on two benchmark datasets: CIFAR-10 \cite{Krizhevsky2009} for image and SNLI \cite{Bowman2015} for text classification. To further examine the applicability and impact of DP-AL in sensitive domains, we employ three medical imaging datasets: BloodCell, a dataset for blood cell classification in microscopy images \cite{Acevedo2020}, RetinalOCT, an imbalanced optical coherence tomography (OCT) dataset of retinal images \cite{Kermany2018}, and CheXpert, a multi-label chest X-ray dataset \cite{Irvin2019}.

The labeling budget is $B = 25,000$ for all datasets, except for BloodCell and SNLI, where the budgets are $B = 2,500$ and $B = 50,000$, respectively.

Note that our experiments simulate a realistic AL scenario, where the limited labeling budget prevents labeling the entire dataset. Thus, the reported metrics are not comparable to scores from other works that are obtained using the entire dataset. However, all values reported in this work employ the same setting, allowing for comparisons between methods and assessment of the methods' utility.

\paragraph{Models} We employ the following model architectures known for their effectiveness under DP-SGD: ResNet-9 with scale normalization \cite{He2016,Klause2022}, Equivariant-ResNet-9 \cite{Hoelzl2023}, and NFNet-F0 \cite{Brock2021,Berrada2023}, with the latter being pretrained on ImageNet \cite{Krizhevsky2012}. The Equivariant ResNet-9 is noteworthy as it comes with optimized hyperparameters for DP-SGD training on CIFAR-10. Therefore, by including both standard and optimized models, we can assess their relative effectiveness under DP-AL. Thus, for CIFAR-10, we compare the performance of both ResNet-9 and Equivariant-ResNet-9. For the BloodCell and RetinalOCT datasets, which are less challenging, we use ResNet-9, while for CheXpert, we employ the more powerful NFNet-F0. SNLI text classification is performed with a pretrained BERT transformer model \cite{Devlin2019,Wolf2020}, where only the final two layers are fine-tuned.

\paragraph{Training Parameters} We adhere to standard training settings from the literature, adjusting only the number of epochs for each specific experiment, as hyperparameter tuning is challenging in realistic AL settings. In line with common practice, we allocate an overall privacy budget of $\eps=8$ and set $\delta = \frac{1}{B}$, where $B$ represents the labeling budget and thus the final dataset size. We conduct four AL iterations ($T=4$), with a privacy budget of $\eps_{\text{Sel}}=2$ for the AL selection, balancing training and selection accuracy. The privacy leakage from DP-SGD training and selection are combined to reach the overall privacy budget as described in \cref{sec:combPLOSS}.

\subsection{Results}

\Cref{tab:performance} presents the results of our main experiments. Note again that the CIFAR-10 results are incomparable with the state-of-the art from literature because they are trained on a subset of the dataset. Below, we highlight the key insights derived from these experiments:

\paragraph{Naive AL vs. Step Amplification (SA) AL} The results demonstrate that methods incorporating step amplification consistently outperform the naive AL method across all datasets. This suggests that our strategy of increasing the number of training steps and fully leveraging the privacy budget of all data points yields strong benefits for DP-AL. For example, on CIFAR-10 (ResNet-9), the step amplification method based on random sampling achieves 66.55\% accuracy, while the naive random approach only achieves 63.40\%. This improvement is mirrored across other datasets, emphasizing the utility of our methodological improvements.

\paragraph{Random AL vs. AL with Acquisition Functions} The introduction of acquisition functions generally leads to improved performance over random sampling in AL, even under DP constraints. Notably, the methods \enquote{minimum margin} and \enquote{entropy} show substantial gains, with both achieving the best performance of AL methods on all datasets and the best overall accuracy on four out of six datasets. Only \enquote{MC BALD} underperforms compared to random selection in several cases. 

These results show that the utility of selection is not fully degraded to random sampling due to privatization. However, the results for non-DP methods reveal that approximately half of the performance benefits of AL are lost due to the effects of privatizing the selection phases. 

\paragraph{Random vs. Random AL} Comparing the two baselines \enquote{Random} (standard DP-SGD training on a randomly selected subset) and \enquote{Random AL} (DP-AL training with random selection) shows that the methodological changes required for DP-AL, \ie, incremental training and iterative data selection, have a limited impact on performance in most experiments. For example, on CIFAR-10 with ResNet-9, the difference is minimal (66.58\% vs. 66.55\%).

However, in other cases, the changes can substantially degrade performance, like on CIFAR-10 with Eq-ResNet-9 (76.45\% vs. 72.23\%) and BloodCell (92.31\% vs. 88.52\%). This drop indicates that, for simpler datasets or when \enquote{naively} transferring architectures which were heavily optimized for training on the full dataset from scratch (\ie, Eq-ResNet-9), the iterative nature of DP-AL may introduce inefficiencies that the informative selection process cannot fully mitigate.

\paragraph{Random vs. AL with Acquisition Functions} Despite the challenges posed by DP, the AL methods can outperform standard DP-SGD training with random sampling (our main baseline) across different modalities, particularly on more complex or unbalanced datasets. For example, on the unbalanced RetinalOCT dataset, the entropy-based AL method achieves a substantially higher accuracy (78.00\%) compared to the random method (74.54\%), underscoring the importance of effective sample selection and identifying underrepresented data points.

\subsection{Ablation Studies}

To better understand the contributions of each individual element in our DP-AL framework, we conduct a series of ablation studies. All experiments in this section focus on the CIFAR-10 dataset with ResNet-9 and utilize entropy sampling as the acquisition function.

\paragraph{Sharing the Privacy Budget}

An important aspect of DP-AL is the shared privacy budget between the selection and training phases, introducing a trade-off between selection and training accuracy. We investigate this trade-off by varying the privacy budget for selection $\epssel$ while keeping the total privacy budget fixed at $\eps = 8$. 

Our results in \cref{fig:budget_split} show that allocating $1 \leq \epssel \leq 3$ leads to improvements over random sampling, with a maximum at $\epssel = 2$. This suggests that a moderate budget for the selection phases strikes a good balance between maximizing the informativeness of selected data points and preserving sufficient privacy budget for effective model training.

\begin{figure}[tb]
    \centering
    \begin{tikzpicture}
        \begin{axis}[
        axis x line=center,
        axis y line=left,
        xmin=-0.01,
        xmax=4,
        ymin=64, 
        ymax=70,
        clip=false,
        xlabel={Selection Privacy Budget $\eps_{\text{Sel}}$},
        ylabel={Accuracy (\%)},
        xtick={0,1,2,3,4},
        legend columns=-1,
        legend style={
            draw=none, fill=none, 
            font=\small,
            at={(1.05,1.3)}
        },
        width=0.97\linewidth,height=4cm,
        domain=6:10]
         \addplot[dashed, thick, Pred] coordinates {(0, 66.58) (4, 66.58)};
        \addplot[dashed, thick, Pblue] coordinates {(0, 68.43) (4, 68.43)}; 
         \addplot [QPblue, mark=*, mark size=1.5pt, error bars/.cd, y dir=both, y explicit] table [x=eps, y=dpal, y error=std] {plot_data/CIFAR10/eps_exp1v2.txt};
        \legend{Random, DP-AL w/o DP selection, DP-AL}
        \end{axis}
    \end{tikzpicture}
    \caption{Performance comparison of entropy sampling on CIFAR-10 for different privacy budgets for selection $\eps_{Sel}$ under an overall privacy budget of $\eps=8$. As baselines we show standard training with a random subset and DP-AL without privatization of the selection phases. We report average $\pm$ standard deviation across 5 runs.}
    \label{fig:budget_split}
\end{figure}

\paragraph{Labeling Budget}

Different labeling budgets/dataset sizes can significantly impact the performance of active learning methods. We investigate this effect by varying the labeling budget $B$ from 15,000 to 45,000 samples on the CIFAR-10 dataset, as shown in \cref{fig:budgets}. 

Our results demonstrate that both AL and random sampling improve in accuracy as the labeling budget increases, which is expected due to the availability of more training data. Interestingly, the performance gap between AL and random sampling varies across different labeling budgets: AL outperforms random sampling at lower budgets (15,000-25,000), suggesting AL's effectiveness with limited data. However, this advantage diminishes as the labeling budget increases, with performances converging at higher budgets (30,000-45,000). These results indicate that our methods are most beneficial when working with smaller datasets, where sample selection has a more significant impact, as seen in privacy-sensitive domains like medicine.

\pgfplotsset{scaled x ticks=false}
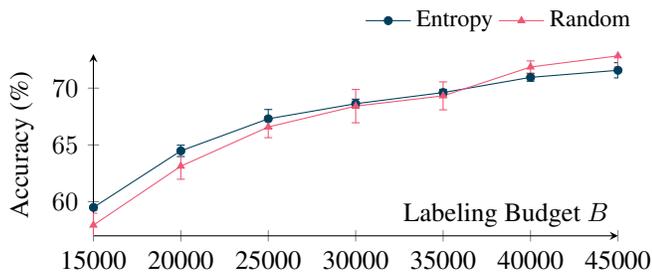
\begin{figure}[tb]
    \centering
    \begin{tikzpicture}
        \begin{axis}[
        axis x line=center,
        axis y line=left,
        xmin=14999,
        xmax=45000,
        xtick={15000,20000,25000,30000,35000,40000, 45000},
        xticklabels={15000,20000,25000,30000,35000,40000,45000},
        ymin=57, 
        ymax=73,
        clip=true,
        xlabel={Labeling Budget $B$},
        ylabel={Accuracy (\%)},
        width=0.97\linewidth,height=4cm,
        legend columns=-1,
        legend style={
            draw=none, fill=none, 
            font=\small,
            at={(1.05,1.3)}
        },
        domain=30:90]
          \addplot [QPblue, mark=*, mark size=1.5pt, error bars/.cd, y dir=both, y explicit] table [x=size, y=entropy, y error=estd] {plot_data/CIFAR10/data_size.txt};
          \addplot [Pred, mark=triangle*, mark size=1.5pt, error bars/.cd, y dir=both, y explicit] table [x=size, y=random, y error=rstd] {plot_data/CIFAR10/data_size.txt};
          \legend{Entropy, Random}
        \end{axis}
    \end{tikzpicture}
    \caption{Performance comparison of standard DP-SGD training with a random subset and AL entropy sampling on CIFAR-10 under different labeling budgets. We report average $\pm$ standard deviation across 5 runs.}
    \label{fig:budgets}
\end{figure}

\paragraph{Overall Privacy Budget}

Finally, in \cref{tab:ovr_priv_budgets}, we evaluate the impact of the overall privacy budget $\eps$. Note that as the overall privacy budget increases, the budget allocated to the selection phases $\epssel$ can also be increased, allowing for more accurate and informative sample selection during AL.

As expected, our results show that higher $\eps$ values lead to improved accuracy for both random sampling and entropy-based AL. This is because a larger privacy budget allows for less noise in the gradient updates during training, thereby improving model convergence and overall performance. Notably, while AL falls short of random sampling at lower privacy budgets, it becomes increasingly advantageous as the privacy budget grows. This suggests that AL is most effective in scenarios with moderate-to-higher privacy budgets (including the commonly used value of $\eps=8$), where the selection process can be more accurate and impactful.

\begin{table}[t]
    \centering
    \caption{Performance comparison of standard DP-SGD training with a random subset and AL entropy sampling on CIFAR-10 under different overall privacy budgets $\eps$. We report average $\pm$ standard deviation across 5 runs.}
    \label{tab:ovr_priv_budgets}
    \begin{tabular}{ccccc}
    \toprule
        & \multicolumn{4}{c}{Accuracy (\%)}\\\midrule
       $\eps$ ($\eps_{\text{Sel}}$)  & 3 (1) & 5 (1) & 8 (2) & 16 (4) \\\midrule
        Random & $57.73_{\pm 0.85}$ & $62.66_{\pm 0.96}$ & $66.58_{\pm 0.94}$ & $70.39_{\pm 0.50}$ \\
        Entropy & $57.42_{\pm 0.91}$ & $63.15_{\pm 0.75}$ & $67.30_{\pm 0.82}$ & $71.93_{\pm 0.48}$ \\
     \bottomrule
    \end{tabular}
\end{table}

\section{Discussion}\label{sec:discussion}

Our work introduces a novel method for privacy-preserving active learning with differential privacy (DP-AL), successfully combining the advantages of active learning's selective data labeling with DP-SGD model training. Through extensive experiments across multiple datasets and model architectures, we have gained several key insights into the viability and challenges of DP-AL.

\paragraph{Conflict Between Data Selection and Membership Inference Accuracy} A fundamental challenge of DP-AL lies in the conflict between AL's goal of selecting \emph{specific} data points and the necessity to protect membership information of those points. AL operates by selecting a subset of data for labeling, essentially making membership decisions about which data points to include in the training set. This decision process mirrors the goal of membership inference attacks (MIAs), where adversaries attempt to infer whether a specific data point was part of the training set. Thus, a deterministic selection process can easily leak membership information.

This creates a direct privacy-utility trade-off: high selection accuracy in AL improves model performance but simultaneously increases the privacy risk through MIAs. As the selection process becomes less random and more targeted, it becomes increasingly susceptible to attacks. Conversely, under stricter privacy constraints, the accuracy of the selection query is reduced, limiting the effectiveness of AL. As a result, achieving both high selection accuracy and strong membership inference protection is inherently difficult.

The implications for weaker threat models or other attacks against data privacy, such as attribute inference and reconstruction attacks, are an interesting topic for future exploration. In particular, it would be interesting to study cases in which membership in the data pool is public, but the actual data point values are not. Ideally, if we could bypass the need for privatization of the selection query, more complex acquisition functions would be viable again and substantial performance improvements would be achievable, as demonstrated by our experiments. While DP-AL outperforms random sampling in specific scenarios, approximately half of AL's performance benefits are lost due to the noise introduced by privatizing the selection process.

\paragraph{Balancing Privacy Budget Allocation} In DP-AL, the privacy budget must be shared between the selection and training phases, adding a new dimension to the privacy-utility trade-off. Our experiments suggest that allocating a moderate portion of the privacy budget to selection ($1 \leq \epssel \leq 3$, from a total $\eps = 8$) can provide a good balance between selecting informative data points and leaving sufficient budget for effective model training. However, small overall privacy budgets leave insufficient room for accurate selection, resulting in no performance improvements over random sampling. 

This trade-off makes DP-AL more complex to implement than standard DP-SGD training, where the entire privacy budget is allocated to model training. The challenge lies in finding the optimal balance between selection accuracy and training effectiveness within the constraints of the overall privacy budget.

\paragraph{Effectiveness of DP-AL on Complex and Imbalanced Datasets} Despite these challenges, DP-AL shows promise in certain domains, particularly when dealing with complex or imbalanced datasets. In medical imaging tasks, such as RetinalOCT, DP-AL significantly outperformed random sampling, with entropy-based AL achieving an accuracy of 78.00\% compared to 74.54\% for random sampling. This improvement is largely due to AL's ability to focus on rare or difficult-to-classify samples, which are essential for enhancing model performance in domains where such data points are crucial.

These results suggest that DP-AL is most beneficial in privacy-sensitive domains where labeled data is both scarce and diverse, as it allows for more efficient use of the labeling budget. However, for simpler or more balanced datasets, such as CIFAR-10, the advantages of DP-AL were less pronounced, and in some instances, the iterative nature of AL introduced inefficiencies, reducing overall performance.

\paragraph{Methodological Contributions} A key methodological contribution of this work is our step amplification method, which optimizes the use of the privacy budget in DP-AL. Step amplification ensures that all data points contribute fully to the model training process, maximizing the utility of each privacy budget. Additionally, by privatizing the selection process and integrating its privacy loss with that of DP-SGD, we further improved the efficiency of privacy management across AL phases. These improvements emphasize the importance of careful allocation and utilization of privacy budgets, particularly in AL settings where the training dataset grows incrementally over time. 

\paragraph{Extension to other ML Methods} While our method is specifically designed for DP-SGD with Poisson sampling in the context of deep neural network training, the core principles of privacy budget allocation and step amplification have the potential to generalize to other ML algorithms and privacy-preserving training techniques. Notably, training other ML models using DP-SGD and incorporating DP-AL is feasible across various model architectures. Moreover, complementary enhancements to DP-SGD, such as denoising techniques \cite{Zhang2024a}, can be seamlessly integrated to further improve performance. An interesting direction for future work would involve exploring alternative batching strategies, such as balls-in-bins sampling \cite{ChoquetteChoo2024}, which could enable integration with methods like DP-FTRL \cite{Kairouz2021} and further expand our method's applicability.

\section{Conclusion}

Our research on differentially private active learning (DP-AL) reveals both promising opportunities and challenges. While DP-AL offers potential benefits in privacy-sensitive domains, especially when dealing with complex or imbalanced datasets, it also faces specific trade-offs due to the inherent conflict between informative sample selection and privacy preservation. This trade-off between selection accuracy and privacy risk implies that DP-AL's effectiveness will likely always be limited under strong privacy constraints. Nonetheless, our research provides insights and methods for integrating AL and DP, and future research should focus on overcoming the limitations discussed here to make active learning more broadly applicable, especially in privacy sensitive domains.

\section*{Acknowledgments}

KS and GK received support from the Bavarian Collaborative Research Project PRIPREKI of the Free State of Bavaria Funding Programme \enquote{Artificial Intelligence -- Data Science}. 

GK received support from the German Federal Ministry of Education and Research and the Bavarian State Ministry for Science and the Arts under the Munich Centre for Machine Learning (MCML), from the German Ministry of Education and Research and the Medical Informatics Initiative as part of the PrivateAIM Project, and from the German Academic Exchange Service (DAAD) under the Kondrad Zuse School of Excellence for Reliable AI (RelAI). 

We thank Julia Moosbauer, Franz Pfister, and Mehmet Yiğit Avcı for the constructive collaboration in the PRIPREKI project leading to this publication.

\bibliographystyle{IEEEtran}
\bibliography{DP}

\twocolumn[\pagebreak]

\appendices

\section{Differential Privacy} \label{apdx:dp}

This section provides mathematical details for some crucial concepts of differential privacy (DP). All these concepts are well-known and do not require any additional proofs. We refer to \cite{Dwork2014, Mironov2017} for details.

\subsection{Rényi Differential Privacy}

Rényi differential privacy (RDP) \cite{Mironov2017} uses the Rényi divergence to quantify privacy loss. For probability distributions $P$ and $Q$, the Rényi divergence of order $\alpha > 1$ is defined as:

\begin{equation}
    \dset_\alpha(P||Q) = \frac{1}{\alpha - 1} \log\mathbb{E}_{x\sim Q}\left(\frac{P(x)}{Q(x)}\right)^\alpha.
\end{equation}

A randomized mechanism $\mech$ satisfies $(\alpha, \eps)$-RDP if for all adjacent datasets $\dset$ and $\dset'$:

\begin{equation}
    \dset_\alpha(\mech(\dset) || \mech(\dset')) \leq \eps.
\end{equation}

RDP offers advantages over $(\eps,\delta)$-DP, including efficient composition theorems, facilitating privacy accounting for complex algorithms. Note that we sometimes use the same symbol $\eps$ for both $(\eps, \delta)$-DP and RDP whenever the exact value of the parameter is irrelevant to the context being discussed.

\subsection{(Additive) Noise Mechanisms}

Noise mechanisms in DP protect the privacy of individuals in a dataset by adding randomness to the outputs of queries or computations. Given a function $f$ applied to dataset $\mathcal{D}$, an (additive) noise mechanism $\mech$ can be defined as:

\begin{equation}
    \mech(f(\mathcal{D})) = f(\mathcal{D}) + \xi,
\end{equation}
where $\xi$ is a random variable sampled from a specific (location/scale-family) distribution.

The noise magnitude required to guarantee DP is determined based on the properties of the function and noise distribution. Typically, the noise scale is adjusted in proportion to the global sensitivity of the function $f$ to establish a specific signal-to-noise ratio. The global sensitivity $\sens{p}$ measures the maximum change in the function's output when a single data point in the dataset is modified. Formally, it is defined as:

\begin{equation}
    \sens{p} = \underset{\mathcal{D}\simeq \mathcal{D}'}{\sup}\left |f(\mathcal{D})-f(\mathcal{D}') \right|_p,
\end{equation}
where $\mathcal{D}\simeq \mathcal{D}'$ indicates that the datasets $\mathcal{D}$ and $\mathcal{D}'$ differ by at most one element (\ie, the datasets are adjacent) and $|\cdot|_p$ denotes a Minkowski $p$-norm.

In the following, we describe two popular noise mechanisms.

\paragraph{Laplace Mechanism}

The Laplace mechanism adds noise to the output of a function $f(\mathcal{D})$ sampled from a Laplace distribution with density:
\begin{equation}
    \text{Lap}(0, \beta) = \frac{1}{2\beta}\exp\left(-\frac{|x|}{\beta}\right),
\end{equation}
where the scale parameter $\beta$ is set in proportion to the global sensitivity $\sens{1}$ based on the $L_1$-norm. 

\begin{theorem}[Laplace mechanism]
The Laplace mechanism guarantees $\eps$-DP for $\beta \geq \sens{1}/\eps$.
\end{theorem}

\paragraph{Gaussian Mechanism}

The Gaussian mechanism adds noise to the output of a function $f(\mathcal{D})$ sampled from a Gaussian distribution:
\begin{equation}
\xi \sim \mathcal{N}(0, \sens{2}^2\sigma^2\mathrm{I}),
\end{equation}
where the noise scale $\sigma$ is set in proportion to the global sensitivity $\sens{2}$ based on the $L_2$-norm.

\begin{theorem}[Gaussian mechanism]
The Gaussian mechanism guarantees ($\eps,\delta$)-DP and $(\alpha, \eps)$-RDP.
\end{theorem}

\subsection{Properties of Differential Privacy}

\begin{theorem}[Post-Processing]\label{thm:post_processing}
    If a mechanism $\mech(\mathcal{D})$ satisfies ($\eps,\delta$)-DP or $(\alpha, \eps)$-RDP, then for any (deterministic or randomized) function $g$, $g(\mech(\mathcal{D}))$ also satisfies ($\eps,\delta$)-DP or $(\alpha, \eps)$-RDP.
\end{theorem}

\begin{theorem}[Basic Sequential Composition]\label{thm:basic_comp}
    If $n$ mechanisms $\mech_i(\mathcal{D})$ satisfy ($\eps_i,\delta_i$)-DP, then the mechanism $\mech'(\mathcal{D}) = (\mech_1(\mathcal{D}), \mech_2(\mathcal{D}), \dots, \mech_n(\mathcal{D}))$ that releases the output of all mechanisms satisfies $\left(\sum_{i=1}^n\eps_i, \sum_{i=1}^n\delta_i\right)$-DP, even if the mechanisms in the sequence are chosen adaptively based on the outputs of the previous mechanism(s).
\end{theorem}

\begin{theorem}[RDP Composition]
    If $n$ mechanisms $\mech_i(\mathcal{D})$ satisfy $(\alpha,\eps_i)$-RDP, then the mechanism $\mech'(\mathcal{D}) = (\mech_1(\mathcal{D}), \mech_2(\mathcal{D}), \dots, \mech_n(\mathcal{D}))$ that releases the output of all mechanisms satisfies $\left(\alpha, \sum_{i=1}^n\eps_i\right)$-RDP, even if the mechanisms in the sequence are chosen adaptively based on the outputs of the previous mechanism(s).
\end{theorem}

\begin{theorem}[Parallel Composition]\label{thm:parallel_comp}
    Let $\mathcal{D}$ be a dataset partitioned into $k$ disjoint subsets $\mathcal{D}_1, \mathcal{D}_2, \dots, \mathcal{D}_k$ and let mechanism $\mech(\mathcal{D})$ satisfy ($\eps,\delta$)-DP. Then, the mechanism $\mech'(\mathcal{D}) = (\mech(\mathcal{D}_1), \mech(\mathcal{D}_2), \dots, \mech(\mathcal{D}_k))$ that releases the outputs of the mechanism applied independently to each subset $\mathcal{D}_i$ satisfies ($\eps,\delta$)-DP.
\end{theorem}

\begin{theorem}[Privacy Amplification by Subsampling]\label{thm:priv_ampl}
    Let $\mech(\mathcal{D})$ be a mechanism that satisfies $(\epsilon, \delta)$-DP. Suppose we construct a new mechanism $\mech'$ by applying $\mech$ to a random subsample of $\mathcal{D}$, where each data point is included in the subsample independently with probability $p<1$. Then, $\mech'$ satisfies $(\epsilon', \delta')$-DP, where $\epsilon' \leq \epsilon$ and $\delta' \leq \delta$.
\end{theorem}

\section{Methodological Details} \label{apdx:met_details}

\subsection{Algorithmic Details} \label{apdx:alg_details}

In \cref{alg:overall}, we summarize our DP-AL method, incorporating step amplification and joint privacy accounting across training and selection phases. Additionally, the auxiliary functions for computing the noise multiplier, sampling rate, and the privacy loss are presented in \cref{alg:getnoise,alg:getsample,alg:geteps}.

\begin{algorithm}[tb]
    \caption{Overall algorithm for DP-AL with step amplification. \texttt{DPSGD-S} is a function that trains a model for a given number of steps using DP-SGD, including batch creation with varying sampling probabilities for different data points. $\lfloor \cdot \rceil$ and $\lfloor \cdot \rfloor$ denote the rounding and floor function, respectively.}
    \label{alg:overall}
	\begin{algorithmic}[1]
        \Require Privacy budget $(\eps, \delta)$, selection privacy budget $\epssel$, selection clip threshold $C_\text{Sel}$, AL uncertainty function $\mathfrak{C}$, initial model parameters $\theta_1$, initial labeled dataset $\mathcal{D}_1$, initial unlabeled dataset $\mathcal{U}_1$, epochs per training phase $e$, batch size $b$, query size $k$, labeling budget $B$
        \State $T \gets \lfloor B/k \rfloor$ \Comment{Num. selection phases}
        \State $q \gets [b/(|\mathcal{D}_1|+(i-1)k)]_{i=1}^{T+1}$ \Comment{Sample rates}
        \State $n \gets [e/q_i]_{i=1}^{T+1}$ \Comment{Training steps}
		\State $\hat{\sigma}\gets \texttt{get\_noiseMultiplier(}\varepsilon,\delta,n,q\texttt{)}$
        \State \textbf{// Step amplification}
        \For{$i = 2, \ldots, T+1$}
        \State $\eps_{\text{target}} \gets \texttt{get\_epsilon(}\delta, \hat{\sigma}, n_{:i}, q_{:i}\texttt{)}$
        \State $\eps_{\text{target,new}} \gets \eps_{\text{target}} - (i-1)\epssel/T$
        \State $q_{\text{old},i} \gets \texttt{get\_sampleRate(}\eps_{\text{target}},\delta,\hat{\sigma},n_{:i}, q_{:{i-1}}\texttt{)}$
        \State $q_{\text{new},i} \gets \texttt{get\_sampleRate}(\eps_{\text{target,new}},\delta,\hat{\sigma},n_{i})$
        \State \textbf{init} $n_{i,\text{low}}\gets n_i$, $n_{i,\text{high}}\gets 3n_i$
        \While{$b \not\approx q_{\text{old},i}|\gset_{\text{old},i}|+q_{\text{new},i}|\gset_{\text{new},i}|$}
        \State $n_i \gets \lfloor (n_{i,\text{high}} - n_{i,\text{low}})/2 \rceil$
        \State $q_{\text{old},i} \gets \texttt{get\_sampleRate(}\eps_{\text{target}},\delta,\hat{\sigma},n_{:i},q_{:{i-1}}\texttt{)}$
        \State $q_{\text{new},i} \gets \texttt{get\_sampleRate}(\eps_{\text{target,new}},\delta,\hat{\sigma},n_{i})$
        \If{$q_{\text{old},i}|\gset_{\text{old},i}|+q_{\text{new},i}|\gset_{\text{new},i}| > b$}
            \State $n_{i,\text{low}} \gets n_i$
        \Else
            \State $n_{i,\text{high}} \gets n_i$
        \EndIf
        \EndWhile
        \EndFor
        \State \textbf{// Iterative DP-AL training}
		\For{$i = 1, \ldots, T$} 
        \State // Training phase
        \State $\theta_{i+1} \gets \texttt{DPSGD-S(}\theta_{i}, \mathcal{D}_{i}, \hat{\sigma}, q_{\text{new},i}, q_{\text{old},i}, n_{i}\texttt{)}$
        \State // Private selection phase
        \State $\xi_\text{Sel} \sim \text{Lap}(0, TC_\text{Sel}/(i\epssel))$ \Comment{Sample noise}
        \State $z_\text{priv} \gets \max(\mathfrak{C}(\mathcal{U},\theta_{i+1}), C_\text{Sel}) + \xi_\text{Sel}$ \Comment{Uncertainties}
		\State $\mathcal{Q}_i \gets {\arg\max}_{\mathcal{S}\subseteq\mathcal{U}, |\mathcal{S}| = k} \sum_{\vect{x}\in \mathcal{S}} z_\text{priv}(\vect{x})$
        \State $\mathcal{D}_{i+1} \gets \mathcal{D}_{i} \cup \mathcal{Q}_i$ \Comment{Labeling}
        \State $\mathcal{U}_{i+1} \gets \mathcal{U}_{i} \setminus \mathcal{Q}_i$ \Comment{Remove from unlabeled set}
		\EndFor
        \State $\theta \gets \texttt{DPSGD-S(}\theta_{T+1}, \mathcal{D}_{T+1}, \hat{\sigma}, q_{\text{new},T+1}, q_{\text{old},T+1}, n_{T+1}\texttt{)}$
	\end{algorithmic}
\end{algorithm}

\begin{algorithm}[tb]
    \caption{Function \texttt{get\_noiseMultiplier}, computing the DP-SGD noise multiplier given target privacy parameters using a binary search algorithm similar to Opacus \cite{Yousefpour2021}.}
    \label{alg:getnoise}
	\begin{algorithmic}[1]
        \Require Target privacy parameters $(\eps, \delta)$, sequence $n$ containing number of training steps $n_i$ for each training phase $i$, sequence $q$ containing sampling rates $q_i$ for each training phase $i$, precision $\gamma=0.01$
        \State \textbf{init} $\eps_\text{high}\gets \infty$
        \State \textbf{init} $\hat{\sigma}_\text{low} \gets 0$, $\hat{\sigma}_\text{high} \gets 10$
        \While {$\eps_\text{high}>\eps$}
            \State $\hat{\sigma}_\text{high} \gets 2\hat{\sigma}_\text{high}$
            \State $\eps_\text{high} \gets \texttt{get\_epsilon(}\delta, \hat{\sigma}_\text{high}, n, q\texttt{)}$
        \EndWhile

        \While{$\eps - \eps_\text{high} > \gamma$}
            \State $\hat{\sigma} \gets (\hat{\sigma}_\text{low}+\hat{\sigma}_\text{high})/2$
            \State $\eps_\text{temp} \gets\texttt{get\_epsilon(}\delta, \hat{\sigma}, n, q\texttt{)}$

            \If{$\eps_\text{temp} < \eps$}
                \State $\hat{\sigma}_\text{high} \gets \hat{\sigma}$
                \State $\eps_\text{high} \gets \eps_\text{temp}$
            \Else
                \State $\hat{\sigma}_\text{low}\gets\hat{\sigma}$
            \EndIf
        \EndWhile
        \State \Return $\hat{\sigma}_\text{high}$
	\end{algorithmic}
\end{algorithm}

\begin{algorithm}[tb]
    \caption{Function \texttt{get\_sampleRate}, which computes the sampling rate for a specific DP-AL training phase using a binary search algorithm.}
    \label{alg:getsample}
	\begin{algorithmic}[1]
        \Require Target privacy parameters $(\eps, \delta)$, noise multiplier $\hat{\sigma}$, current training phase $I$, sequence $n$ containing number of training steps $n_i$ for each training phase $i \leq J$, sequence $q$ containing sampling rates $q_i$ for each previous training phase $i < I$, precision $\gamma=0.001$
        \State \textbf{init} $q_{I,\text{low}} \gets 1\mathrm{e}{-9}$, $q_{I,\text{high}} \gets 1$

        \While{$q_{I,\text{low}} / q_{I,\text{high}} < 1-\gamma$}
            \State $q_I \gets (q_{I,\text{low}}+q_{I,\text{high}})/2$
            \State $q_\text{temp} \gets [q , q_I]$ \Comment{Append $q_I$ to sequence $q$}
            \State $\eps_\text{temp} \gets \texttt{get\_epsilon(}\delta, \hat{\sigma}, n, q_\text{temp}\texttt{)}$

            \If{$\eps_\text{temp} < \eps$}
                \State $q_{I,\text{low}} \gets q_I$
            \Else
                \State $q_{I,\text{high}}\gets q_I$
            \EndIf
        \EndWhile
        \State \Return $q_{I,\text{high}}$
	\end{algorithmic}
\end{algorithm}

\begin{algorithm}[tb]
    \caption{Function \texttt{get\_epsilon}, which computes the privacy loss $\eps$ across DP-AL training phases using heterogeneous composition and the RDP accountant \cite{Mironov2017}. \texttt{compute\_rdp} is a function that numerically computes RDP for the sampled Gaussian mechanism at order $\alpha$ \cite{Yousefpour2021}. For simplicity, finding the optimal $\alpha$ is omitted.}
    \label{alg:geteps}
	\begin{algorithmic}[1]
        \Require Target privacy parameter $\delta$, noise multiplier $\hat{\sigma}$, sequence $n$ containing number of training steps $n_i$ for each training phase $i$, sequence $q$ containing sampling rates $q_i$ for each training phase $i$, number of training phases $I$, optimal RDP parameter $\alpha$
        \For{$i = 1, \ldots, I$}
        \State $\rho_i \gets \texttt{compute\_rdp}(n_i, q_i, \hat{\sigma}, \alpha)$
        \EndFor
        \State $\rho \gets \sum_{i=1}^{I}\rho_i$ \Comment{RDP Composition}
        \State $\eps \gets \rho + \log\frac{\alpha-1}{\alpha} -\frac{\log\delta+\log\alpha}{\alpha-1}$ \Comment{$(\eps,\delta)$-Conversion \cite{Balle2020b}}
        \State \Return $\eps$
	\end{algorithmic}
\end{algorithm}

\section{Experimental details} \label{apdx:exp_details}

\subsection{Dataset details}

The \emph{CIFAR-10} dataset \cite{Krizhevsky2009} is a widely-used benchmark for image classification tasks, consisting of 60,000 $32\stimes32$ color images of natural scenes, divided into 10 classes, with 6,000 images per class. The dataset is split into 50,000 training and 10,000 test images. For validation, we randomly select 10,000 images from the training set.

\emph{BloodCell} \cite{Acevedo2020} is a medical imaging dataset for classification of blood cell types. It includes 8 classes, representing various normal and abnormal blood cell types, such as red blood cells, white blood cells, and platelets, as well as cells indicative of certain blood disorders. The dataset contains a total of 17,092 RGB images split into 11,959 training, 1,712 validation, and 3,421 test images. All images are center-cropped and resized to $128\stimes128$.

The \emph{RetinalOCT} \cite{Kermany2018} dataset is designed for the classification of retinal diseases using optical coherence tomography (OCT) scans. The dataset consists of 4 classes, representing various retinal conditions, including choroidal neovascularization, diabetic macular edema, drusen, and normal retinal tissue. It contains a total of 109,309 grayscale images, with 97,477 training images, 10,832 validation images, and 1,000 test images. All images are center-cropped and resized to $128\stimes128$. The training and validation datasets are imbalanced, with a significant variation in the number of images per class, while the test dataset is balanced.

The \emph{CheXpert} \cite{Irvin2019} dataset is a large-scale dataset for medical multi-label classification, containing chest X-ray images along with corresponding clinical reports and metadata. It includes 14 classes, representing various conditions, such as pneumonia, pleural effusion, and cardiomegaly, as well as a class for no finding. The dataset contains a total of 224,316 grayscale images from 65,240 patients split into 223,582 training, 234 validation, and 500 test images. All images are center-cropped and resized to $192\stimes192$. For simplicity, following \cite{Berrada2023}, we assume that each image is captured from a unique patient. Our training utilizes all 14 classes, while evaluation focuses on a subset of 5 key conditions: \enquote{Atelectasis}, \enquote{Cardiomegaly}, \enquote{Consolidation}, \enquote{Edema}, and \enquote{Pleural Effusion}. Uncertain labels in the dataset are handled using label smoothing \cite{Szegedy2016} with value of 0.2. Here, for the classes \enquote{Atelectasis}, \enquote{Edema}, and \enquote{Pleural Effusion}, uncertain labels are mapped to the positive label and for the remaining classes to the negative label.

The \emph{Stanford Natural Language Inference} (SNLI) \cite{Bowman2015} dataset, is a large-scale benchmark for evaluating natural language understanding models. It contains 570,000 sentence pairs, each labeled with one of three categories: entailment, contradiction, or neutral. The set is split into 550,152 training, 10,000 validation, and 10,000 test pairs.

\subsection{Frameworks}

Our implementation employs \texttt{PyTorch} \cite{Paszke2019} for core machine learning functionalities and \texttt{Opacus} \cite{Yousefpour2021} for privacy-preserving techniques. For image data, we employ CIFAR-10 from the \texttt{torchvision} library and the RetinalOCT and BloodCell datasets from the \texttt{MedMnistv2} collection \cite{Yang2023c}. We utilize a pretrained BERT model sourced from the \texttt{transformers} library \cite{Wolf2020}. All other models and datasets are obtained from their respective original implementations.

\subsection{Hyperparameters}

\begin{table*}[tb]
    \centering
    \caption{Hyperparameters. The column \enquote{Query sizes} denotes the number of labeled data points in every AL selection phase (the first value is the size of the randomly sampled initial labeled dataset). \enquote{SGD-AGC} denotes DP-SGD with adaptive gradient clipping \cite{Brock2021}.}
    \label{tab:exp_details}
    \begin{tabular}{ccccccccc}
        \toprule
       Dataset  & Image size & Labeling budget $B$ & Query sizes $Q$ & Batch & Epochs (AL) & LR & Clip & Optimizer  \\ \midrule
        CIFAR-10 & $32\stimes 32\stimes 3$ & 25,000 (50\%) & [10,000, 10,000, 3,000, 1,000, 1,000] & 4096 & 100 (150) & 0.001 & 1.0 & nAdam  \\
        CIFAR-10 (Eq) & $32\stimes 32\stimes 3$ & 25,000 (50\%) & [10,000, 10,000, 3,000, 1,000, 1,000] & 4096 & 300 (300) & 2.0 & 2.0 & SGD \\
        BloodCell & $128\stimes 128 \stimes 3$ & 2,500 ($\approx$25\%) & [1024, 1024, 300, 100, 52] & 512 & 90 (105) & 0.002 & 1.0 & nAdam \\
        RetinalOCT & $128\stimes 128\stimes 1$ & 25,000 ($\approx$25\%) & [10,000, 10,000, 3,000, 1,000, 1,000]  & 4096 & 100 (150) & 0.001 & 1.0 & nAdam \\
        CheXpert & $192\stimes 192\stimes 1$ & 25,000 ($\approx$10\%) & [10,000, 10,000, 3,000, 1,000, 1,000] & 4096 & 100 (150) & 1.0 & 1.0 & SGD-AGC\\
        SNLI & - & 50,000 ($\approx$10\%) & [20,000, 20,000, 6,000, 2,000, 2,000] & 4096 & 50 (75) & 0.001 & 1.0 & AdamW  \\
        \bottomrule 
    \end{tabular}
\end{table*}

\paragraph{Example Figures}
For \cref{fig:eps_impl,fig:steps_probs,fig:eps_impl2}, we assume a DP-AL process with a total privacy budget of $\eps=8$, $\delta=0.0004$, noise multiplier $\hat{\sigma}=4.08$, number of epochs $e=30$, initial labeled dataset size $N=10,000$, initial unlabeled dataset size $M=50,000$, labeling budget $B=25,000$, query size $Q=3,750$, and expected batch size $b=4096$.

\paragraph{Training Runs}

\Cref{tab:exp_details} summarizes the hyperparameters used for our main experiments.

\section{Active Learning Acquisition Functions}\label{apdx:acFunctions}

In this section, we describe the active learning (AL) acquisition functions used in our experiments. We also analyze their global sensitivity, which is required for privatizing the selection query with differential privacy (DP).

\subsection{Least Confidence \cite{Culotta2005} }

The least confidence method selects the $k$ data points with the lowest confidence, \ie, the lowest predicted probability for the most likely class. For a given data point $\vect{x}$, the confidence is defined as

\begin{equation}
F(\vect{x}) = \max_{c} p(c \given \vect{x}; \theta),
\end{equation}

where $p(c \given \vect{x}; \theta)$ represents the predicted probability of class $c$ given the data point $\vect{x}$ and model parameters $\theta$.

In the case of a uniform probability distribution, where $p(c \given \vect{x}; \theta) = 1/C~\forall c\in C$, the model is completely uncertain and the lowest possible predicted probability is achieved. Thus, the range of confidence values is $[1/C, 1]$ and the global sensitivity of the least confidence measure is $\Delta(F) = 1-1/C$.

In the context of multi-label classification, confidence can be calculated for each label independently using the sigmoid outputs. Since each label is binary, the lowest possible predicted probability is 0.5, and $\Delta(F) = 1/2$.

\subsection{Minimum Margin \cite{Scheffer2001}}

In minimum margin sampling, data points that are closest to the decision boundary are selected. This is determined by computing the smallest margin between the highest and second-highest predicted class probabilities as follows:

\begin{equation}
M(\vect{x}) = \hat{p}_1(\vect{x}; \theta) - \hat{p}_2(\vect{x}; \theta),
\end{equation}
where $\hat{p}_1(\vect{x}; \theta)$ and $\hat{p}_2(\vect{x}; \theta)$ are the highest and second-highest class probabilities predicted by the model, respectively.

The margin values range between $[0, 1]$. Thus, the global sensitivity is $\Delta(M) = 1$.

Note that this approach does not naturally extend to multi-label classification scenarios. Thus, we excluded it from our (multi-label) CheXpert experiments.

\subsection{Entropy \cite{Joshi2009}}

Entropy sampling involves labeling the $k$ data points with the highest discrete normalized entropy of class posterior probabilities. The entropy for a data point $\vect{x}$ is defined as follows:

\begin{equation}
	H(\vect{x}) = -\dfrac{\sum_{c=1}^{C}p(c\given \vect{x};\theta)\log_2p(c\given \vect{x};\theta)}{\log_2C},
\end{equation}
where $C$ denotes the number of classes.

For multi-label classification, we average the entropies of individual class probabilities obtained from the sigmoid function.

The value range of the normalized entropy is $[0,1]$. Thus, the global sensitivity is $\Delta(H) = 1$. However, since neural networks are usually overconfident, we find that setting $\Delta(\hat{H}) = 0.8$ by clipping ($\hat{H} = \min(H, 0.8)$) improves utility when adding DP noise.

\subsection{BALD \cite{Houlsby2011} with Monte Carlo Dropout \cite{Gal2017}}

Bayesian Active Learning by Disagreement (BALD) is an acquisition function that leverages Bayesian neural networks to quantify uncertainty. To approximate Bayesian inference without training a full Bayesian network, Monte Carlo (MC) dropout can be employed. This involves performing multiple stochastic forward passes through the network with dropout layers active during inference, effectively creating an ensemble of models.

The core idea behind BALD is to select samples for which the models/forward passes exhibit high disagreement. This is achieved by computing the mutual information between the model output and the model parameters, which is approximated as the difference between the entropy of the average output and the average entropy of individual outputs:

\begin{equation}
    I(\vect{x}) \approx \tilde{H}(\vect{x}) - \frac{1}{J}\sum_{j=1}^JH_j(\vect{x}),
\end{equation}
where $C$ is the number of classes, $J$ is the number of models (\ie, forward passes), and

\begin{equation}
    \tilde{H}(\vect{x}) = -\dfrac{\sum_{c=1}^{C}\frac{1}{J}\sum_{j=0}^J p_j(c\given \vect{x};\theta)\log_2\frac{1}{J}\sum_{j=0}^J p_j(c\given \vect{x};\theta)}{\log_2C}.
\end{equation}

The value range lies again between 0 and 1. Thus, the global sensitivity is $\Delta(I) = 1$. However, since a large disagreement between models is highly unlikely and the mutual information is rather small, we set the sensitivity to $\Delta(\hat{I}) = 0.5$ by clipping: $\hat{I} = \min(I, 0.5)$. This improves utility when adding DP noise.

\section{Additional Experiments}

\subsection{Runtime Analysis of Step Amplification}\label{apdx:runtime}

We empirically evaluate the runtime overhead of step amplification. Step amplification is a preprocessing step that occurs once before training. Our analysis varies relevant parameters for step amplification (labeling budget, query sizes, batch size, and number of epochs) by using the parameters from CIFAR-10, BloodCell, and SNLI experiments (see \cref{tab:exp_details}). Additionally, we vary the number of planned training phases. All experiments are conducted on an AMD Ryzen Threadripper PRO 5975WX 32-Cores CPU.

The results in \cref{tab:runtime} demonstrate that step amplification introduces only a small runtime increase to DP-AL of a few minutes across all experimental settings. Variations in experimental parameters demonstrate negligible impact on runtime, while the number of training phases emerges as the primary computational cost driver, with runtime scaling approximately linearly as phases increase.

We acknowledge that more sophisticated search algorithms or complex privacy accountants could potentially increase runtime. However, the current implementation demonstrates minimal overhead that does not compromise the overall DP-AL workflow.

\begin{table}[tb]
    \centering
    \caption{Runtime of step amplification (in seconds) across different parameter configurations and numbers of training phases. Average $\pm$ standard deviation across 10 runs.}
    \label{tab:runtime}
    \begin{tabular}{cccc}
    \toprule
      Training phases & CIFAR-10 & BloodCell & SNLI \\\midrule
       2  & $37.8_{\pm 0.2}$ & $28.2_{\pm 0.1}$ & $29.9_{\pm 0.2}$ \\
       3  & $67.1_{\pm 0.3}$ & $54.0_{\pm 2.1}$ & $56.7_{\pm 0.1}$ \\
       4  & $110.9_{\pm 1.1}$ & $104.8_{\pm 0.6}$ & $87.5_{\pm 0.3}$\\
       5  & $137.2_{\pm 0.6}$ & $127.5_{\pm 0.9}$ & $126.3_{\pm 1.4}$ \\
       6  & $163.3_{\pm 2.5}$ & $158.5_{\pm 1.2}$ & $142.9_{\pm 2.0}$ \\\bottomrule
    \end{tabular}
\end{table}

\subsection{Alternatives to Step Amplification}\label{apdx:dispLeakage}

As discussed in the main text, we proposed step amplification to achieve a different privacy leakage between groups of data points. This approach leverages the fact that newly labeled data have not participated in any training steps, thus retaining a higher privacy budget for the remaining training (assuming a lower privacy leakage from selection). In this experiment, we explore alternatives to step amplification. Specifically, we consider a \enquote{continual learning} setting, where the model is first trained exclusively on the new data until all points have obtained the same privacy loss and \enquote{noise reduction}, where sampling probabilities are increased as in step amplification, but only the noise multiplier is adjusted to maintain the same expected batch size.

The results in \cref{tab:disp_leakage} show that continual learning performs the worst, likely due to catastrophic forgetting as the model temporarily focuses solely on new data. Noise reduction and step amplification show similar performance, with step amplification maintaining a slight accuracy advantage.

\begin{table}[tb]
    \centering
    \caption{Performance comparison of methods for handling disparate privacy leakage in DP-AL. Experiments are conducted on CIFAR-10 with ResNet-9. Results are reported as mean $\pm$ standard deviation over 5 training runs. The best performance is highlighted in bold.}
    \label{tab:disp_leakage}
    \begin{tabular}{cc}
    \toprule
      Method & Accuracy (\%)  \\\midrule
       Continual learning  & $63.94_{\pm 0.98}$ \\
       Noise reduction  & $66.83_{\pm 0.72}$ \\
       Step amplification  & \bm{$67.30_{\pm 0.82}$} \\\bottomrule
    \end{tabular}
\end{table}

\subsection{Uncertainty Distribution}

\begin{figure*}[tb]
    \centering
    \begin{tikzpicture}
        \begin{axis}[
            axis x line=center,
            axis y line=left,
            ybar,
            bar shift=0pt,
            clip=false,
            xlabel={Entropy},
            title={Phase 1},
            title style = {font=\small, at={(0.05,0.95)}},
            width=.49\linewidth,height=3.5cm,
            xmin=-0.04,
            xmax=1.04,
            ymin=0,
            scaled y ticks=false,
            y filter/.expression={y==0 ? nan : y},
            legend columns=-1,
            xlabel style={at={(ticklabel cs:0.5)}, below=0pt},
        legend style={
            draw=none, fill=none, 
            font=\small,
            at={(1.05,1.3)}
        },
        ]
        \addplot table [x=bins, y=countsEnt] {plot_data/CIFAR10/hist1.txt};
        \addplot table [x=bins, y=countsTrue] {plot_data/CIFAR10/hist1.txt};
        \end{axis}
        \end{tikzpicture}
    \hfill
        \begin{tikzpicture}
        \begin{axis}[
            axis x line=center,
            axis y line=left,
            clip=false,
            ybar,
            bar shift=0pt,
            xlabel={Entropy},
            title={Phase 2},
            title style = {font=\small, at={(0.05,0.95)}},
            xlabel style={at={(ticklabel cs:0.5)}, below=0pt},
            width=.49\linewidth,height=3.5cm,
            y filter/.expression={y==0 ? nan : y},
            xmin=-0.04,
            xmax=1.04,
            ymin=0,
            scaled y ticks=false,
            legend columns=-1,
        legend style={
            draw=none, fill=none, 
            font=\small,
            at={(1.05,1.3)}
        },
        ]
        \addplot table [x=bins, y=countsEnt] {plot_data/CIFAR10/hist2.txt};
        \addplot table [x=bins, y=countsTrue] {plot_data/CIFAR10/hist2.txt};
        \end{axis}
        \end{tikzpicture}\\
        \begin{tikzpicture}
        \begin{axis}[
            axis x line=center,
            axis y line=left,
            clip=false,
            ybar,
            bar shift=0pt,
            xlabel={Entropy},
            title={Phase 3},
            title style = {font=\small, at={(0.05,0.95)}},
            xlabel style={at={(ticklabel cs:0.5)}, below=0pt},
            width=.49\linewidth,height=3.5cm,
            xmin=-0.04,
            scaled y ticks=false,
            y filter/.expression={y==0 ? nan : y},
            xmax=1.04,
            legend columns=-1,
        legend style={
            draw=none, fill=none, 
            font=\small,
            at={(1.05,1.3)}
        },
        ]
        \addplot table [x=bins, y=countsEnt] {plot_data/CIFAR10/hist3.txt};
        \addplot table [x=bins, y=countsTrue] {plot_data/CIFAR10/hist3.txt};
        \end{axis}
        \end{tikzpicture}
    \hfill
        \begin{tikzpicture}
        \begin{axis}[
            axis x line=center,
            axis y line=left,
            clip=false,
            ybar,
            scaled y ticks=false,
            bar shift=0pt,
            xlabel={Entropy},
            y filter/.expression={y==0 ? nan : y},
            title={Phase 4},
            title style = {font=\small, at={(0.05,0.95)}},
            xlabel style={at={(ticklabel cs:0.5)}, below=0pt},
            width=.49\linewidth,height=3.5cm,
            xmin=-0.04,
            ymin=0,
            xmax=1.04,
            legend columns=-1,
        legend style={
            draw=none, fill=none, 
            font=\small,
            at={(1.05,1.3)}
        },
        ]
        \addplot table [x=bins, y=countsEnt] {plot_data/CIFAR10/hist4.txt};
        \addplot table [x=bins, y=countsTrue] {plot_data/CIFAR10/hist4.txt};
        \end{axis}
        \end{tikzpicture}
    \caption{Empirical distribution of entropy values across the four AL selection phases (left to right). The top-$k$ entropies corresponding to the data points that should be selected in each phase, are highlighted in \emph{red}. The results originate from our CIFAR-10 ResNet-9 experiment, where the query size changes across selection phases.}
    \label{fig:hist_entropy}
\end{figure*}
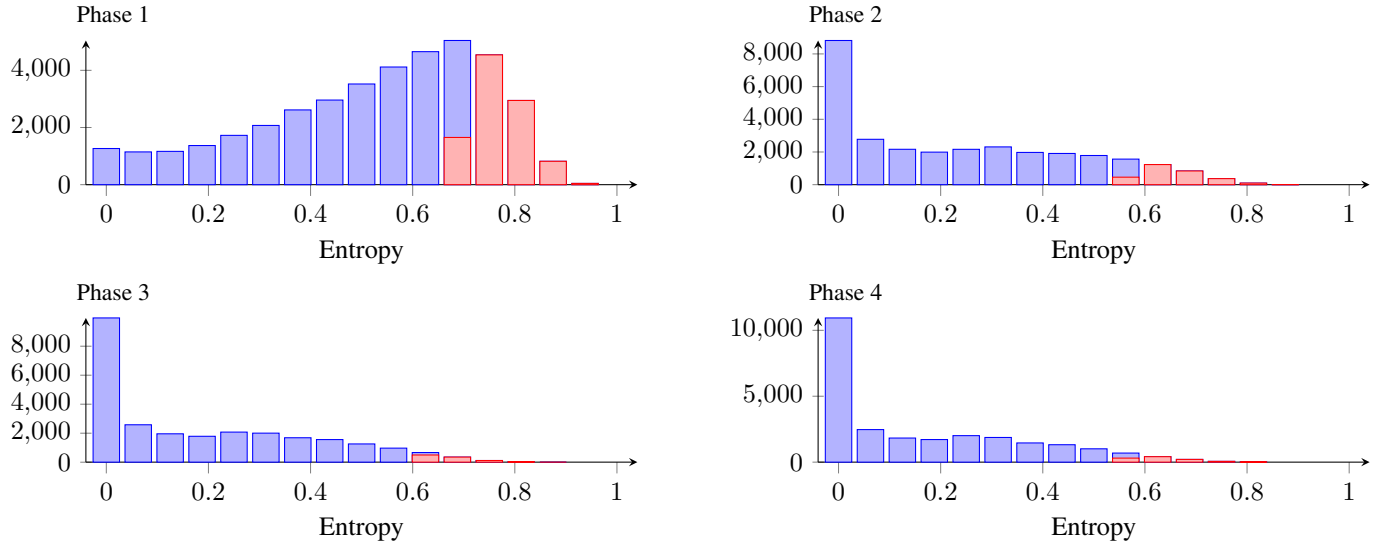

\Cref{fig:hist_entropy} illustrates the empirical distribution of uncertainty (entropy) values across the four active learning (AL) selection phases in our CIFAR-10 ResNet-9 experiment. The results show that the distribution of uncertainties evolves substantially across phases. In Phase 1, the uncertainty distribution is slightly skewed towards higher values. As training progresses, we observe a significant shift towards lower uncertainty values. By Phase 4, the majority of samples exhibit very low entropy, indicating the model's increased confidence. 

In each phase, the red-highlighted areas represent the top-$k$ entropies, corresponding to the \enquote{ideal} data points for selection. Notably, these high-entropy samples become increasingly rare in later phases, underscoring the importance of accurate selection as training advances.

\subsection{Selection Accuracy}

\begin{table}[tb]
    \centering
    \caption{Relative improvement in selection accuracy between random selection and AL entropy selection privatized with $\eps_{\text{Sel}}=2$ across different selection phases. Note that the results originate from our CIFAR-10 ResNet-9 experiment, where the query size changes across selection phases.}
    \label{tab:sel_accs}
    \begin{tabular}{ccccc}
    \toprule
       Phase & Metric & Random & DP-AL & Rel. improvement \\\midrule
        & Accuracy & $24.28\%$ & $29.17\%$ & $20.14\%$ \\
        1 & IoU & 0.138 & 0.171 & $23.91\%$\\
        & MSE & 0.115 & 0.096 & $16.52\%$ \\\midrule
        & Accuracy & $8.63\%$ & $13.16\%$ & $52.49\% $\\
        2 & IoU & 0.045 & 0.071 & $57.78\%$\\
        & MSE & 0.237 & 0.211 & $10.97\%$ \\\midrule
        & Accuracy & $2.90\%$ & $5.40\%$ & $86.21\%$ \\
        3 & IoU & 0.015 & 0.028 & $86.67\%$\\
        & MSE & 0.279 & 0.259 & $7.17\%$\\ \midrule
        & Accuracy & $3.00\%$ & $5.30\%$ & $76.67\%$ \\
        4 & IoU & 0.015 & 0.027 & $80.00\%$\\
        & MSE & 0.269 & 0.244 & $9.29\%$\\
        \bottomrule
    \end{tabular}
\end{table}

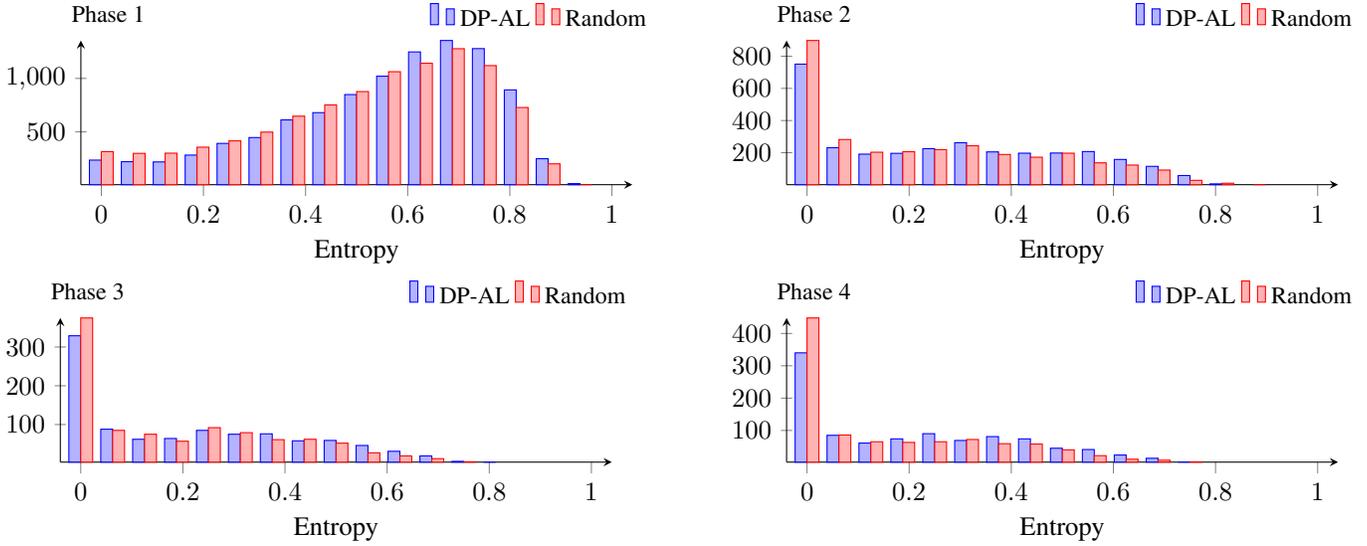
\begin{figure*}[tb]
    \centering
    \begin{tikzpicture}
        \begin{axis}[
            axis x line=center,
            axis y line=left,
            clip=false,
            ybar,
            bar width=4.5pt,
            bar shift=0pt,
            xlabel={Entropy},
            title={Phase 1},
            title style = {font=\small, at={(0.05,0.95)}},
            width=.49\linewidth,height=3.5cm,
            xmin=-0.04,
            xmax=1.04,
            y filter/.expression={y==0 ? nan : y},
            scaled y ticks=false,
            legend columns=-1,
            xlabel style={at={(ticklabel cs:0.5)}, below=0pt},
        legend style={
            draw=none, fill=none, 
            font=\small,
            at={(1.05,1.3)}
        },
        ]
        \addplot+[bar shift=-2.25pt] table [x=bins, y=countsSel] {plot_data/CIFAR10/hist1.txt};
        \addplot+[bar shift=2.25pt] table [x=bins, y=countsRand] {plot_data/CIFAR10/hist1.txt};
        \legend{DP-AL, Random}
        \end{axis}
        \end{tikzpicture}
    \hfill
        \begin{tikzpicture}
        \begin{axis}[
            axis x line=center,
            axis y line=left,
            clip=false,
            ybar,
            bar width=4.5pt,
            bar shift=0pt,
            title={Phase 2},
            title style = {font=\small, at={(0.05,0.95)}},
            xlabel={Entropy},
            xlabel style={at={(ticklabel cs:0.5)}, below=0pt},
            width=.49\linewidth,height=3.5cm,
            xmin=-0.04,
            xmax=1.04,
            y filter/.expression={y==0 ? nan : y},
            scaled y ticks=false,
            legend columns=-1,
        legend style={
            draw=none, fill=none, 
            font=\small,
            at={(1.05,1.3)}
        },
        ]
        \addplot+[bar shift=-2.25pt] table [x=bins, y=countsSel] {plot_data/CIFAR10/hist2.txt};
        \addplot+[bar shift=2.25pt] table [x=bins, y=countsRand] {plot_data/CIFAR10/hist2.txt};
        \legend{DP-AL, Random}
        \end{axis}
        \end{tikzpicture}\\
        \begin{tikzpicture}
        \begin{axis}[
            axis x line=center,
            axis y line=left,
            clip=false,
            ybar,
            y filter/.expression={y==0 ? nan : y},
            bar width=4.5pt,
            bar shift=0pt,
            title={Phase 3},
            title style = {font=\small, at={(0.05,0.95)}},
            xlabel={Entropy},
            xlabel style={at={(ticklabel cs:0.5)}, below=0pt},
            width=.49\linewidth,height=3.5cm,
            xmin=-0.04,
            scaled y ticks=false,
            xmax=1.04,
            legend columns=-1,
        legend style={
            draw=none, fill=none, 
            font=\small,
            at={(1.05,1.3)}
        },
        ]
        \addplot+[bar shift=-2.25pt] table [x=bins, y=countsSel] {plot_data/CIFAR10/hist3.txt};
        \addplot+[bar shift=2.25pt] table [x=bins, y=countsRand] {plot_data/CIFAR10/hist3.txt};
        \legend{DP-AL, Random}
        \end{axis}
        \end{tikzpicture}
    \hfill
        \begin{tikzpicture}
        \begin{axis}[
            axis x line=center,
            axis y line=left,
            clip=false,
            ybar,
            y filter/.expression={y==0 ? nan : y},
            scaled y ticks=false,
            bar shift=0pt,
            title={Phase 4},
            title style = {font=\small, at={(0.05,0.95)}},
            xlabel={Entropy},
            xlabel style={at={(ticklabel cs:0.5)}, below=0pt},
            width=.49\linewidth,height=3.5cm,
            xmin=-0.04,
            bar width=4.5pt,
            xmax=1.04,
            legend columns=-1,
        legend style={
            draw=none, fill=none, 
            font=\small,
            at={(1.05,1.3)}
        },
        ]
        \addplot+[bar shift=-2.25pt] table [x=bins, y=countsSel] {plot_data/CIFAR10/hist4.txt};
        \addplot+[bar shift=2.25pt] table [x=bins, y=countsRand] {plot_data/CIFAR10/hist4.txt};
        \legend{DP-AL, Random}
        \end{axis}
        \end{tikzpicture}
    \caption{Empirical distribution of \emph{selected} entropy values across four selection phases (left to right). The entire AL selection process was privatized by $\eps_{\text{Sel}} = 2$ using the Laplace mechanism. The results originate from our CIFAR-10 ResNet-9 experiment, where the query size changes across selection phases.}
    \label{fig:hist_entropy2}
\end{figure*}

\Cref{fig:hist_entropy2} demonstrates the effectiveness of our DP-AL approach in selecting high-uncertainty samples compared to random sampling. In Phase 1, DP-AL shows a modest advantage, which becomes more pronounced in subsequent phases. Overall, DP-AL consistently selects samples with higher entropy values compared to random sampling. 

\Cref{tab:sel_accs} quantifies this performance advantage. DP-AL consistently outperforms random sampling across all phases, with improvements in accuracy, Intersection over Union (IoU), and Mean Squared Error (MSE). Notably, the relative improvement in accuracy increases from approximately 20\% in Phase 1 to around 80\% in Phases 3 and 4, highlighting DP-AL's growing advantage as high-entropy samples become rarer.

However, DP-AL's distribution differs significantly from the \enquote{ideal} selection (highlighted red in \cref{fig:hist_entropy}), due to the privacy constraints ($\eps_{\text{Sel}}$ = 2 using the Laplace mechanism). Overall, despite these limitations, DP-AL offers consistent improvements over random sampling in identifying informative samples throughout the training process.

\subsection{Noise Mechanisms for Private Selection} \label{apdx:sel_mechanisms}

\begin{figure*}[tb]
    \centering
\begin{tikzpicture}
        \begin{axis}[
        axis x line=center,
        axis y line=left,
        xmin=1,
        xmax=10,
        ymin=20, 
        ymax=50,
        clip=false,
        xlabel={Selection Privacy Budget $\eps_{\text{Sel}}$},
        xlabel style={at={(ticklabel cs:0.5)}, below=0pt},
        ylabel={Accuracy (\%)},
        legend columns=2,
        legend style={
            draw=none, fill=none, 
            font=\small,
            at={(1.05,1.3)}
        },
        width=0.85\linewidth,height=6.cm,
        domain=1:10]
         \addplot [QPblue, mark=*, mark size=1.5pt, error bars/.cd, y dir=both, y explicit] table [x=eps, y=lapl, y error=laplstd] {plot_data/CIFAR10/sel_accs.txt};
         \addplot [Pred, mark=triangle*, mark size=1.5pt, error bars/.cd, y dir=both, y explicit] table [x=eps, y=laplC] {plot_data/CIFAR10/sel_accs.txt};
         \addplot [Plila, mark=*, mark size=1.5pt, error bars/.cd, y dir=both, y explicit] table [x=eps, y=gaus, y error=gausstd] {plot_data/CIFAR10/sel_accs.txt};
         \addplot [Porange, mark=triangle*, mark size=1.5pt, error bars/.cd, y dir=both, y explicit] table [x=eps, y=subgaus, y error=subgausstd] {plot_data/CIFAR10/sel_accs.txt};
         \addplot [Pgreen, mark=*, mark size=1.5pt, error bars/.cd, y dir=both, y explicit] table [x=eps, y=rand, y error=randstd] {plot_data/CIFAR10/sel_accs.txt};
        \legend{Laplace, Laplace ($\sens{}=0.8$), Gaussian, Gaussian ($p=0.7$), Random}
        \end{axis}
    \end{tikzpicture}
    \caption{Accuracy of the AL selection privatized with different noise mechanisms and privacy budgets. Next to the standard mechanisms, we consider the Laplace mechanism with clipping uncertainties to $0.8$ (effectively reducing sensitivity) and the subsampled Gaussian mechanism with $p=0.7$. The source distribution originates from the first selection phase of our CIFAR-10 ResNet-9 experiment with entropy sampling.}
    \label{fig:mechaccs}
\end{figure*}
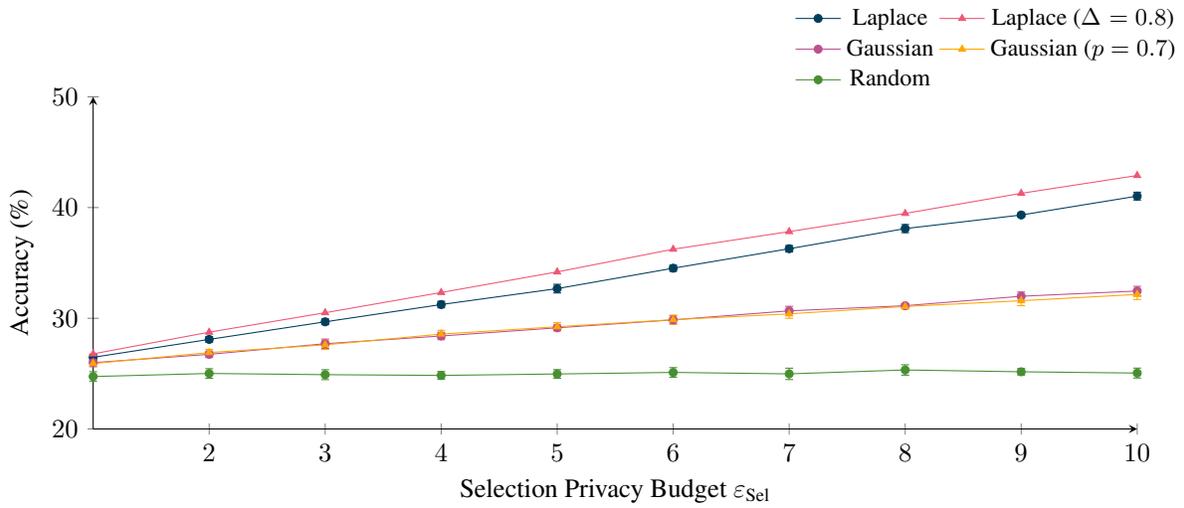

\Cref{fig:mechaccs} compares various noise mechanisms for privatizing the AL selection across different privacy budgets $\eps_{\text{Sel}}$. The Laplace mechanism consistently outperforms other methods, with a modified version using our clipping trick ($\sens{} = 0.8$) showing slight improvements, especially at higher privacy budgets. This suggests that optimizing the sensitivity can yield benefits. Interestingly, the subsampled Gaussian mechanism does not improve upon the standard Gaussian approach, indicating that subsampling may not be well-suited for our selection setting.

Overall, all mechanisms show improvements over random selection, with the gap widening as the privacy budget increases. This demonstrates that even under strict privacy constraints, AL can provide benefits over random sampling.

\end{document}